\documentclass[10pt,twocolumn,letterpaper]{article}
\PassOptionsToPackage{table}{xcolor}
\usepackage[pagenumbers]{cvpr}
\usepackage[utf8]{inputenc}
\usepackage[T1]{fontenc}
\usepackage{mathtools}
\newcounter{proposition}
\newenvironment{proposition}[1][]{\par\addvspace{\medskipamount}\noindent\refstepcounter{proposition}\textbf{Proposition~\theproposition}\if\relax\detokenize{#1}\relax\else\ (#1)\fi\textbf{.}\ \itshape\ignorespaces}{\par\addvspace{\medskipamount}}
\newenvironment{proof}{\par\noindent\textit{Proof.}\ \ignorespaces}{\hfill$\square$\par\addvspace{\medskipamount}}
\newcounter{definition}
\newenvironment{definition}[1][]{\par\addvspace{\medskipamount}\noindent\refstepcounter{definition}\textbf{Definition~\thedefinition}\if\relax\detokenize{#1}\relax\else\ (#1)\fi\textbf{.}\ \ignorespaces}{\par\addvspace{\medskipamount}}
\usepackage{multirow}
\usepackage{makecell}
\usepackage{fontawesome5}
\newsavebox{\benchmarkbox}
\newcommand{\benchmarktable}[3]{%
  \begingroup
  \setlength{\tabcolsep}{3pt}%
  \sbox{\benchmarkbox}{\begin{tabular}{#1}#3\end{tabular}}%
  \addtolength{\tabcolsep}{\dimexpr(\linewidth-\wd\benchmarkbox)/\numexpr2*#2\relax\relax}%
  \begin{tabular}{#1}#3\end{tabular}%
  \endgroup
}
\usepackage{tikz}
\usetikzlibrary{arrows.meta,calc,decorations.markings,shadows}
\usepackage{nicefrac}
\usepackage{microtype}
\usepackage{stfloats}
\usepackage{siunitx}
\usepackage{soul}
\definecolor{cvprblue}{rgb}{0.21,0.49,0.74}
\usepackage[pagebackref,breaklinks,colorlinks,allcolors=cvprblue]{hyperref}
\hypersetup{pdftitle={Generation as Reverse Encoding},
  pdfauthor={Cheng Shi, Sibei Yang, Hanzhong Guo, Yipeng Qin, Pietro Lio, Yizhou Yu}}

\def\confName{CVPR}
\def\confYear{2027}

\definecolor{myblue}{RGB}{50,100,200}

\DeclareRobustCommand{\revadd}[1]{{\color{black}#1}}

\title{Generation as Reverse Encoding}
\author{%
  Cheng Shi\textsuperscript{1}\quad Sibei Yang\textsuperscript{2,\,\faIcon[regular]{envelope}}\quad Hanzhong Guo\textsuperscript{1}\quad Yipeng Qin\textsuperscript{3}\quad Pietro Lio\textsuperscript{4}\quad Yizhou Yu\textsuperscript{1,\,\faIcon[regular]{envelope}}\\[3pt]
  {\normalsize \textsuperscript{1}University of Hong Kong\quad
   \textsuperscript{2}Sun Yat-sen University\quad
   \textsuperscript{3}Cardiff University\quad
   \textsuperscript{4}University of Cambridge}\\
  {\small\hypersetup{urlcolor=black}\href{mailto:shicheng2025@connect.hku.hk}{\texttt{shicheng2025@connect.hku.hk}},
   \href{mailto:yangsb3@mail.sysu.edu.cn}{\texttt{yangsb3@mail.sysu.edu.cn}},
   \href{mailto:yizhouy@acm.org}{\texttt{yizhouy@acm.org}}}\\
  {\small \faIcon[regular]{envelope}\enspace Corresponding authors.}%
}

\makeatletter
\newcommand{\maketitlewithteaser}{%
  \twocolumn[{\@twocolumnfalse\maketitle\begin{minipage}{\textwidth}
    \centering

        \centering
        \includegraphics[width=\linewidth]{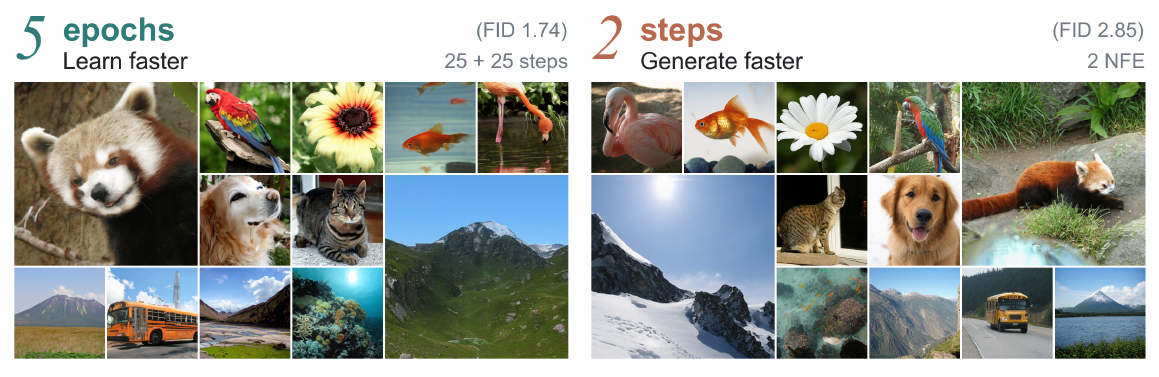}
    \captionsetup{hypcap=false}\captionof{figure}{\textbf{RevFlow samples on ImageNet 256$\times$256.}
\textbf{Left: learn faster.} After only 5 training epochs, RevFlow-H reaches 1.74 FID with 25 sampling steps per stage.
\textbf{Right: generate faster.} After native representation matching, RevFlow-XL reaches 2.85 FID with only 2 NFE in total (one per stage).
Both follow the reverse-encoding path: generating from class labels to deep representations, then to shallow representations, before finally decoding to pixels.}
    \label{fig:teaser}
\end{minipage}
\vspace{12pt}}]%
}
\makeatother

\begin{document}
\maketitlewithteaser

\begin{abstract}

Pretrained visual representations accelerate generative learning by abstracting away \textcolor{black}{appearance and spatial detail}. Yet the same abstraction makes the return to pixels far harder, shifting much of the burden onto a larger decoder.
We ask why \textbf{the representation-to-pixel path} should be learned independently when the encoder \textbf{already provides a learned pixel-to-semantics path (the semantic encoding path)}. We introduce \textbf{RevFlow}, based on the principle of \textbf{generation as reverse encoding.}
\textcolor{black}{RevFlow uses intermediate representations of a frozen encoder as successive generative targets, first generating deep semantic representations, then shallow representations that retain appearance and spatial detail, before decoding to pixels.} \textcolor{black}{To learn these transitions without training a separate generator for each stage, we develop \textbf{HyC-DiT}, a unified generative model with a standard DiT backbone shared across stages and stage-specific condition injection.} \textcolor{black}{We further show that native matching in semantic representation space enables few-step generation.}
RevFlow accelerates both learning and sampling. On ImageNet \(256\times256\), RevFlow reaches 1.96 FID at 690M and 1.74 FID at 1.55B after only 5 training epochs. It also achieves 30.25 dB reconstruction PSNR with an 89M decoder. \textcolor{black}{After a short native-matching adaptation phase}, RevFlow-XL achieves 2.85 FID using a single sampling step in each stage. Our code repository is at \href{https://github.com/ChengShiest/RevFlow}{\nolinkurl{github.com/ChengShiest/RevFlow}}.

\end{abstract}

\section{Introduction}

\begin{figure*}[t]
    \centering
    \includegraphics[width=\linewidth]{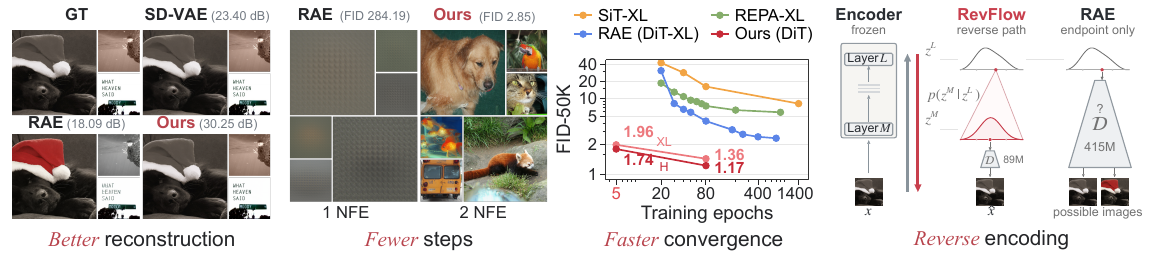}

\caption{
\textbf{Reconstruction fidelity, few-step generation, fast convergence, and reverse encoding.}
{\textbf{Left:} ImageNet validation reconstructions from SD-VAE, RAE, and RevFlow, with PSNR in parentheses and decoder sizes of 49M, 415M, and 89M. RAE changes the gray hat to red; RevFlow preserves its color.}
{\textbf{Middle:} RAE without matching (1 NFE) and RevFlow-XL with native matching (2 NFE, one per stage). Table~\ref{tab:one_step} reports results without CFG for both adaptations.}
{\textbf{Right:} ImageNet convergence of RevFlow-XL (690M) and RevFlow-H (1.55B) with standard DiT backbones.}
{\textbf{Far right:} RAE generates the encoder's endpoint, while RevFlow uses the semantic encoding path to generate deep and then shallow representations before pixel decoding.}
}
    \label{fig:intro}
\end{figure*}

{Visual encoding and image reconstruction/generation can be viewed as opposite processes.} {A pretrained visual encoder progressively transforms an image into increasingly abstract representations, whereas image reconstruction/generation maps such representations back to concrete visual realizations.}
In conventional autoencoding~\cite{yao2025vavae,razavi2019vqvae2}, the reverse problem is posed as reconstruction: a representation is paired with the particular image that produced it, and the decoder is trained to recover that image.
\textit{\textbf{Generation poses a harder reverse problem.}} Once abstraction makes different images share similar representations, the inverse becomes underdetermined: a representation can correspond to multiple images rather than a fixed target.

This distinction becomes especially pronounced for representations pretrained for visual understanding\footnote{In this paper, unless otherwise specified, \emph{encoder} refers to a frozen pretrained visual representation encoder, such as DINO or CLIP, rather than to a reconstruction-oriented VAE encoder.}. {Reconstruction-oriented representations, such as VAE latents, retain appearance and spatial detail for faithful recovery, whereas semantic representations learned for visual understanding suppress variations less relevant to semantic discrimination.} Such abstraction accelerates generative learning, as exploited by REPA~\citep{yu2025repa} and RAE~\citep{zheng2025rae}, {but leaves more appearance and spatial detail to be generated on the way back to pixels.} In RAE-style models, this burden is delegated to a separately learned decoder, which is typically around 10$\times$ larger than those in conventional latent diffusion~\citep{rombach2022ldm}; Figure~\ref{fig:intro} further shows that RAE can reconstruct a gray Christmas hat as red.
Abstraction therefore shifts, rather than removes, the difficulty: \emph{{generation in a semantic representation space is easier to learn}, while decoding becomes more demanding and absorbs an increasing share of generative modeling.}

{Recent methods mitigate this tension by making the generated representation retain more appearance and spatial detail.} VA-VAE~\citep{yao2025vavae} enriches reconstruction-oriented latents with semantic structure, while RAEv2~\citep{singh2026raev2} {combines multiple encoder levels into a single target containing both semantic and spatial information.}
{These methods improve the target through alignment or feature aggregation, then learn a separate decoding path from that target to pixels.}

We take a different view. Here we define the \emph{semantic encoding path} as the progression of representations from pixels to abstract semantics that semantic pretraining forms within an encoder. \emph{\textbf{If generation ultimately reverses semantic encoding, why learn a separate decoding path back to pixels when the semantic encoding path already connects pixels to semantics?}} We propose to reverse this path itself. {Along the semantic encoding path, shallow representations retain more appearance and spatial detail, while deep representations emphasize semantics.} The path thus acts like the forward process of diffusion models, which removes information gradually so that generation can recover it through simple reverse steps~\citep{sohl2015deep,ho2020ddpm}. Unlike data-agnostic Gaussian corruption, which removes information roughly by spatial frequency~\citep{rissanen2023ihdm}, the encoder removes it in a learned order: appearance and spatial detail first, semantics last~\citep{mahendran2015understanding,achille2018invariance}. Its reverse resolves semantics first, from deep representations that largely reveal the class, leaving representation learning, a main bottleneck of diffusion training~\citep{yu2025repa}, to the pretrained encoder. Directly predicting clean shallow features from a class label faces an uninformative conditional mean and wide residual variance, forcing stiff jumps onto dispersed clean Dirac targets rather than smooth score refinement. Reversing the path keeps training smooth: the deep stage first models a lower-variance semantic space, and its output then provides spatial mean guidance that leaves only a narrow residual variance for the shallow stage (Appendix~\ref{sec:theory}, Sec.~\ref{sec:ablations}). This suggests a different principle: \emph{{intermediate representations along the semantic encoding path should serve as successive generation targets, \textbf{defining how generation unfolds}}.} Generating each target, from the class label or from the next deeper representation on the path, forms a \emph{stage}. How to learn stages that differ in conditions and representation spaces within one model, however, remains open. %

We propose \textbf{RevFlow} (\emph{\textbf{generation as reverse encoding}}), which carries out multiple stages of generation along the reversed semantic encoding path with a single Transformer backbone. The generation targets of all stages are representations of the same image along the path. Generation proceeds from class labels to deep semantic representations, then shallow representations, and finally pixels. %
{Sharing a backbone across stages requires accommodating different conditions and addressing optimization interference across representation spaces.} We therefore introduce \textbf{HyC-DiT}, which adapts a standard DiT~\cite{peebles2023dit} in both architecture and training: {global class conditioning is used in the first stage and spatial feature conditioning in the second, while clean-target prediction enables effective weight sharing across stages.}
{Crucially, the deep-to-shallow stage models appearance and spatial detail before pixel decoding, reducing the generative modeling burden on the decoder. The complete pipeline} {pairs a lightweight decoder with better generation and reconstruction.}

\textcolor{black}{Beyond reverse encoding, we study native distribution matching for few-step adaptation of semantic representation generators.} Existing methods such as FD-loss~\citep{yang2026fdloss} and representation distribution matching~\citep{feng2026rdm} improve few-step generation by aligning generated and real samples in pretrained feature space, but conventional VAE-based generators {typically need to decode} latents into images and re-encode them for matching. \textcolor{black}{Generating encoder representations allows native matching without this decode--encode loop. Native matching (NM) improves both single-stage RAE and two-stage RevFlow, demonstrating applicability beyond our reverse-encoding architecture.}

On class-conditional ImageNet 256$\times$256, RevFlow with a standard DiT backbone reaches 1.96 FID with {690M} parameters and 1.74 FID with 1.55B parameters after \textbf{only 5 training epochs}. For reconstruction, RevFlow achieves 30.25 dB PSNR with an 89M decoder, compared with 18.09 dB using the 415M RAE decoder. \textcolor{black}{After a short native-matching adaptation phase}, RevFlow-XL reaches 2.85 FID using one sampling step per stage and \textcolor{black}{no CFG}.

Our contributions are:
\begin{itemize}[leftmargin=1.5em,itemsep=1pt,topsep=2pt]
\item We introduce \emph{generation as reverse encoding}, {modeling the inverse of representation encoding as a generative process that reverses the semantic encoding path of a frozen encoder from deep to shallow representations.} %
\item We develop HyC-DiT, a unified generative model that learns all stages of the reversed path with one shared Transformer backbone. It injects the class label as a global condition and the deep representation as a spatial condition through per-token AdaLN. Clean-target prediction avoids the degradation that velocity prediction suffers under sharing, matching separate generators with about half the parameters.
\item \textcolor{black}{We show that pretrained semantic representations can serve as both generation targets and native distribution-matching spaces. Native matching enables few-step adaptation in both single-stage RAE and two-stage RevFlow without pixel decoding or re-encoding.}
\item {We provide a fast-converging training recipe that achieves 1.96/1.74 FID with standard DiT backbones on ImageNet $256\times256$ after only five epochs, making high-quality generative modeling more accessible for rapid experimentation.}
\end{itemize}

\section{Related Work}
\label{sec:related}

\noindent\textbf{Target spaces for generation.}
Diffusion and flow models~\citep{ho2020ddpm,song2021score,lipman2023flow,liu2023flow,albergo2023stochastic,peebles2023dit,ma2024sit} are trained either in pixel space~\citep{adm,hoogeboom2023simple,li2025jit} or in the latents of reconstruction-trained tokenizers~\citep{kingma2014vae,vandenoord2017vqvae,esser2021taming,rombach2022ldm}. Recent work brings in semantics from pretrained encoders: VA-VAE~\citep{yao2025vavae} and REPA~\citep{yu2025repa} align VAE latents or diffusion hidden states with frozen encoder features, REPA-E~\citep{repa-e} tunes the VAE end to end, and other tokenizers are designed for generation~\citep{yang2025detok,yue2025uniflow}. RAE~\citep{zheng2025rae} generates final-layer encoder features directly, and RAEv2~\citep{singh2026raev2} fuses several layers into one latent. All of them generate a single target and leave the rest to a separately learned decoder, which absorbs more of the synthesis as the target becomes more abstract. RevFlow instead distributes generation across successive levels of the encoder hierarchy.

\noindent\textbf{Inverting visual representations.}
Feature inversion~\citep{mahendran2015understanding,dosovitskiy2016inverting} shows that deeper layers become increasingly invariant, retaining semantics while losing appearance and spatial detail; pretrained Vision Transformers~\citep{caron2021dino,oquab2024dinov2,simeoni2025dinov3,he2022mae} follow a similar trend. In our ImageNet retrieval examples, shallow features match color and texture, while deep features more often retrieve images of the same class. Such inversions, like autoencoder and RAE decoders, are deterministic regressions. Because encoding is many-to-one, RevFlow models each inversion step generatively and uses deterministic pixel decoding only at a shallow level that retains image detail.

\noindent\textbf{Multi-stage and representation-conditioned generation.}
Coarse-to-fine generators decompose synthesis across resolutions~\citep{karras2018progressive,razavi2019vqvae2,ho2022cascaded,gu2024matryoshka,VAR}. Others first generate a compact representation before synthesizing the image: unCLIP~\citep{ramesh2022unclip} samples CLIP embeddings, RCG~\citep{li2024rcg} generates self-supervised representations, W"urstchen~\citep{pernias2024wuerstchen} generates a compressed semantic latent, and Latent Forcing~\citep{baade2026latentforcing} denoises features ahead of pixels. In these methods, the representation conditions generation in another space; in RevFlow, successive targets are paired features from a single frozen encoder, each stage reverses one segment of the encoding path, and all stages share a common backbone. For spatial conditioning, HyC-DiT follows spatially adaptive normalization~\citep{park2019spade} rather than using auxiliary control branches~\citep{zhang2023controlnet}, computing adaLN-Zero parameters on a per-token basis.

\noindent\textbf{Few-step generation and distribution matching.}
Few-step generators can be obtained through distillation~\citep{salimans2022progressive,yin2024dmd}, consistency training~\citep{song2023consistency}, or average-velocity modeling~\citep{geng2025meanflow}. Generators can also be trained by matching distributions in fixed feature spaces, from maximum mean discrepancy~\citep{gretton2012mmd,li2015gmmn} to recent objectives defined on pretrained representations~\citep{yang2026fdloss,feng2026rdm,deng2026drifting}. For latent-space generators, such representation-based matching requires decoding generated latents to pixels and re-encoding them into the matching space; RevFlow instead generates encoder representations directly and matches them natively.

\section{RevFlow: Generation as Reverse Encoding}
\label{sec:method}
\begin{figure*}[t]
    \centering

    \begin{tikzpicture}
    \node[anchor=south west,inner sep=0] (frameworkimage) at (0,0)
      {\includegraphics[width=\textwidth]{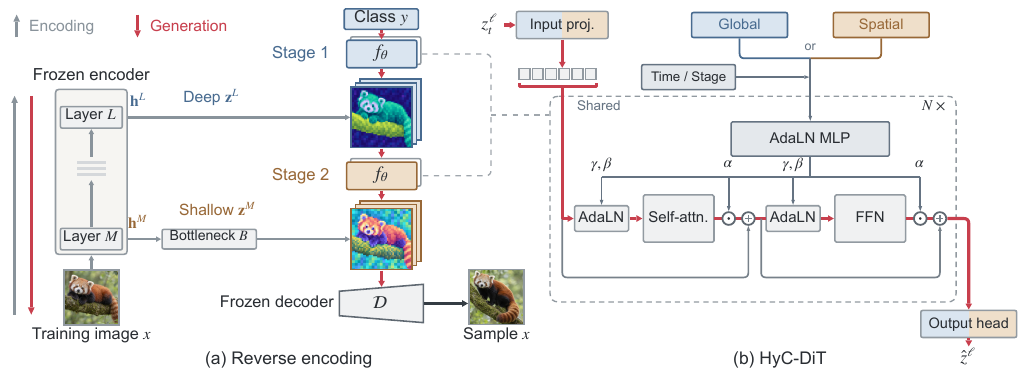}};
    \begin{scope}[x={(frameworkimage.south east)},y={(frameworkimage.north west)}]
      \definecolor{targetprojbg}{RGB}{240,241,242}
      \definecolor{outputprojleft}{RGB}{218,229,240}
      \definecolor{outputprojright}{RGB}{239,223,202}
      \fill[targetprojbg] (0.1580,0.3355) rectangle (0.2477,0.3837);
      \node[anchor=center,inner sep=0] at (0.20286,0.35918)
        {\resizebox{0.085\textwidth}{!}{{\sffamily Projection $P_{\mathrm{target}}$}}};
      \fill[outputprojleft] (0.898,0.108) rectangle (0.939394,0.157);
      \fill[outputprojright] (0.939394,0.108) rectangle (0.983,0.157);
      \node[anchor=center,inner sep=0] at (0.939394,0.132)
        {\resizebox{0.078\textwidth}{!}{{\sffamily Output proj.}}};
    \end{scope}
    \end{tikzpicture}
    \caption{\textbf{Reverse encoding with HyC-DiT.}
\textbf{(a) Reverse encoding.}
{A frozen encoder defines the semantic encoding path from shallow to deep representations. RevFlow reverses this path:} Stage~1 generates the deep target $\boldsymbol{z}^{L}$ from class $y$, Stage~2 generates the shallow target $\boldsymbol{z}^{M}$ conditioned on $\boldsymbol{z}^{L}$, followed by pixel decoding. {$P_{\mathrm{target}}$ maps shallow encoder features to the Stage~2 target.} \textcolor{black}{Gray and red arrows denote encoding and generation, respectively.}
\textbf{(b) HyC-DiT.}
{The Transformer blocks are shared across both stages, while the target input/output projections and condition mappings are stage-specific.}
 Stage~1 uses global class conditioning, whereas Stage~2 uses token-wise spatial conditioning from the aligned deep representation. The conditions modulate AdaLN shift and scale $(\beta,\gamma)$ and residual gates $\alpha$ in each block.
}
\label{fig:framework}
\end{figure*}

A frozen semantic encoder maps an image $\boldsymbol{x}$ to representations $\boldsymbol{h}^{\ell}=\mathcal{E}^{\ell}(\boldsymbol{x})$ along the semantic encoding path. Deep representations emphasize semantics, while shallow representations retain more appearance and spatial detail. RevFlow divides the reversed path into multiple stages, generating deep and then shallow representations before pixel decoding (Figure~\ref{fig:framework}a).
{We address three design questions: which representation levels define the generation targets (Sec.~\ref{sec:path}); how a unified model shares its Transformer backbone across stages through suitable conditioning and prediction parameterization (Secs.~\ref{sec:architecture}--\ref{sec:prediction_space}); and how native matching enables one-step sampling per stage (Sec.~\ref{sec:one_step}).}

\subsection{Constructing the Reverse-encoding Path}
\label{sec:path}

\noindent\textbf{\revadd{Mean guidance and variance reduction.}}
Unlike diffusion, which reverses a fixed corruption~\citep{rissanen2023ihdm,bansal2023cold}, RevFlow reverses the semantic encoding path, whose order of information removal is formed by semantic pretraining.
In clean-target flow matching on $\boldsymbol{a}_{t}=(1-t)\boldsymbol{a}+t\boldsymbol{\epsilon}$, decomposing a clean target $\boldsymbol{a}=\mathbb{E}[\boldsymbol{a}|\boldsymbol{u}]+\boldsymbol{r}$ separates mean regression $\mathbb{E}[\boldsymbol{a}|\boldsymbol{u}]$ from residual denoising $\mathbb{E}[\boldsymbol{r}|\boldsymbol{a}_{t},\boldsymbol{u}]$, whose input gradient is $\frac{1-t}{t^{2}}\operatorname{Cov}(\boldsymbol{r}|\boldsymbol{a}_{t},\boldsymbol{u})$ (Prop.~\ref{prop:flow_variance} in Appendix~\ref{sec:theory}). Directly predicting shallow representation $\boldsymbol{h}^{M}$ from class label $y$ faces an uninformative mean and wide residual variance $V(\boldsymbol{h}^{M}|y)$, averaging out detail at high noise and forcing stiff Dirac jumps as $t\to 0$. Reversing the path keeps training smooth: Stage~1 first predicts deep representation $\boldsymbol{h}^{L}$ in a lower-variance semantic space, and $\boldsymbol{h}^{L}$ then supplies spatial mean guidance $\mathbb{E}[\boldsymbol{h}^{M}|\boldsymbol{h}^{L}]$ so Stage~2 refines a narrow residual $V(\boldsymbol{h}^{M}|\boldsymbol{h}^{L})$ before low-variance pixel decoding~\citep{blau2018perception} (Fig.~\ref{fig:intro}).

\begin{figure}[t]
    \centering
    \begin{tikzpicture}
    \node[anchor=south west,inner sep=0] (layerimage) at (0,0)
      {\includegraphics[width=\linewidth]{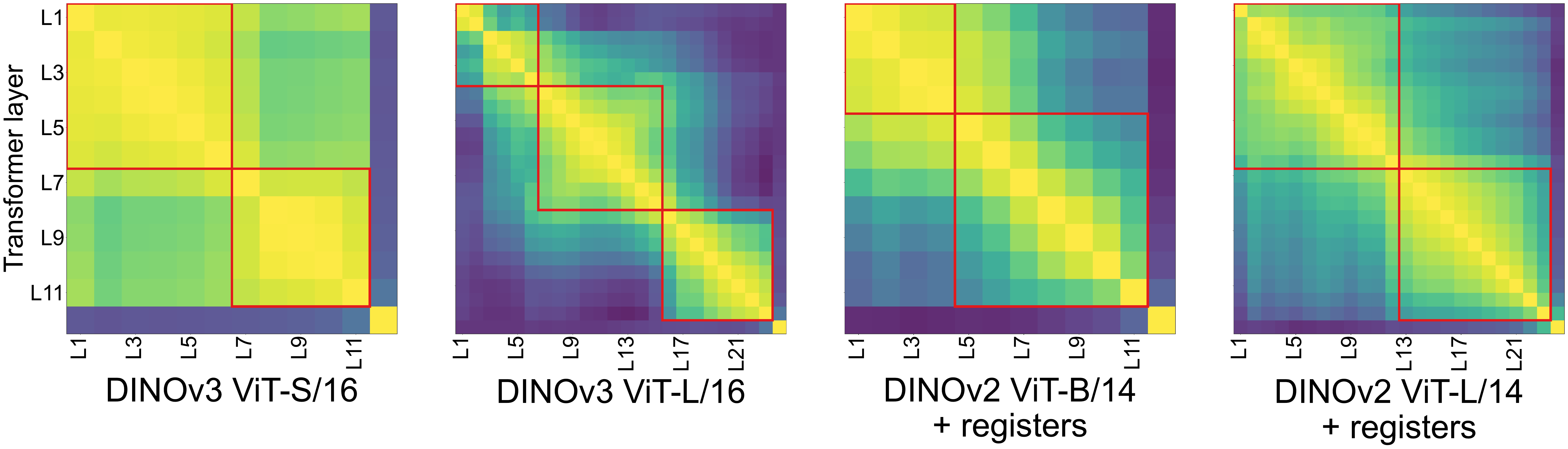}};
    \begin{scope}[x={(layerimage.south east)},y={(layerimage.north west)}]
      \fill[white] (0.3135,0.075) rectangle (0.4780,0.180);
      \fill[white] (0.5610,0.000) rectangle (0.7270,0.182);
      \node[anchor=base west,inner sep=0] at (0.316453,0.102048)
        {\resizebox{0.159228\linewidth}{!}{{\sffamily DINOv3 ViT-L/16}}};
      \node[anchor=base west,inner sep=0] at (0.563403,0.105597)
        {{\setlength{\unitlength}{\linewidth}\begin{picture}(0,0)
          \put(0,0){\resizebox{0.161573\linewidth}{!}{\sffamily DINOv2 ViT-B/14}}
          \put(0.031186,-0.022502){\resizebox{0.099172\linewidth}{!}{\sffamily + registers}}
        \end{picture}}};
    \end{scope}
    \end{tikzpicture}
    \caption{\textbf{Layer-wise similarity in frozen encoders.}
    Each entry reports the cosine similarity between CLS representations from two transformer layers, averaged over images. Consecutive layers form high-similarity regimes separated by clear transitions. RevFlow selects one representative generation target from each regime rather than traversing every encoder layer.}
    \label{fig:layer_similarity}
\end{figure}

\noindent\textbf{Selecting abstraction levels.}
A ViT encoder contains many layers, but traversing every layer would substantially increase sampling cost and accumulate \revadd{upstream mean-guidance drift} across stages. We therefore select a small set of representative abstraction levels. {In Figure~\ref{fig:layer_similarity}, consecutive layers form high-similarity regimes separated by clear transitions, so the semantic encoding path can be summarized by a few abstraction levels.}

Across the encoders we examine, the similarity structure is dominated by an earlier and a later regime. We therefore instantiate RevFlow with two generative stages ($K=2$, $\ell_{2}=L$, $\ell_{1}=M$). {We select the deep representation $\boldsymbol{h}^{L}$ from the later regime for its semantic structure, and the shallow representation $\boldsymbol{h}^{M}$ from the earlier regime for its appearance and spatial detail.} For the DINOv3-L encoder used in our main models, we take $\boldsymbol{h}^{M}$ from layer~7, at the end of the early regime, and $\boldsymbol{h}^{L}$ from layer~23, at the end of the late regime. Layer indices are zero-based. Layer~23 is the output of the final Transformer block, taken before the encoder's final normalization.

\noindent\textbf{Instantiating the reverse path.}
{RevFlow generates an image through}
\begin{equation}
    y\rightarrow\boldsymbol{z}^{L}\rightarrow\boldsymbol{z}^{M}\rightarrow\boldsymbol{x}.
    \label{eq:reverse_path}
\end{equation}
{Here, $\boldsymbol{z}^{L}=\boldsymbol{h}^{L}$ and $\boldsymbol{z}^{M}=P_{\mathrm{target}}(\boldsymbol{h}^{M})$, where $P_{\mathrm{target}}$ projects the shallow representation into a compact target space while preserving its spatial layout. Following RAE~\citep{zheng2025rae}, the target projection and pixel decoder are trained jointly for reconstruction with the encoder frozen, then fixed during generative training.} The projection reduces the shallow feature from 1024 to 128 channels on a $16\times16$ grid. A decompressor restores the channel dimension before a 12-block Transformer decoder maps the grid to pixels.

{For each training image, the encoder and target projection provide paired targets $(\boldsymbol{z}^{L},\boldsymbol{z}^{M})$. Stage~1 generates $\boldsymbol{z}^{L}$ from the class condition, and Stage~2 generates $\boldsymbol{z}^{M}$ conditioned on $\boldsymbol{z}^{L}$. The final shallow representation is decoded into pixels.}

\subsection{HyC-DiT Architecture: Hybrid Conditioning}
\label{sec:architecture}

\noindent\textbf{Global and spatial conditioning.}
{HyC-DiT implements both stages within a unified generative model using a standard DiT backbone~\citep{peebles2023dit}. The backbone is shared across stages, while the conditioning pathway accepts a global class embedding in Stage~1 and spatially aligned tokens in Stage~2. The latter preserves correspondence with the shallow target, which would be lost by collapsing the condition into a single vector.}
For each stage $\ell\in\{L,M\}$, we sample $t\sim p(t)$ and
$\boldsymbol{\epsilon}^{\ell}\sim\mathcal{N}(\boldsymbol{0},\mathbf{I})$, and form
\begin{equation}
    \boldsymbol{z}^{\ell}_{t}
    =(1-t)\boldsymbol{z}^{\ell}
    +t\boldsymbol{\epsilon}^{\ell}.
    \label{eq:interpolation}
\end{equation}
Both stages use $f_{\theta}(\cdot)$ to predict clean targets $\hat{\boldsymbol{z}}^{\ell}$:
\begin{equation}
    \hat{\boldsymbol{z}}^{\ell}
    =f_{\theta}
    (\boldsymbol{z}^{\ell}_{t},t,\boldsymbol{c}^{\ell}),
    \qquad
    \boldsymbol{c}^{L}=y,
    \qquad \boldsymbol{c}^{M}=\tilde{\boldsymbol{z}}^{L}
    \label{eq:stage_predictions}
\end{equation}
where $\tilde{\boldsymbol{z}}^{L}$ is the deep-feature condition for Stage~2.

\noindent\textbf{Condition noise.}
Because Stage~2 is trained with deep representations extracted from real images but sampled with those generated by Stage~1, we perturb the training condition as
\begin{equation}
    \tilde{\boldsymbol{z}}^{L}
    =
    \sqrt{1-\sigma}\,\boldsymbol{z}^{L}
    +
    \sqrt{\sigma}\,\boldsymbol{\epsilon}_{c},
    \qquad
    \boldsymbol{\epsilon}_{c}\sim
    \mathcal{N}(\boldsymbol{0},\mathbf{I}),
    \label{eq:condition_noise}
\end{equation}
where $\sigma\sim\mathrm{Uniform}[\sigma_{\min},\sigma_{\max}]$.
The noise level $\sigma$ is also provided to the model{, mirroring noise-conditioning augmentation in cascaded diffusion~\citep{ho2022cascaded}}.

\noindent\textbf{Global and spatial AdaLN.}
Let $T$ denote the timestep embedding, $E_y$ the class embedding, $E_s$ a learned stage embedding, $E_\sigma$ the condition-noise embedding, {and $P_{\mathrm{cond}}$ a linear projection that maps the deep condition to the Transformer width.} The two stages form
\begin{equation}
\begin{aligned}
    \boldsymbol{m}^{L}
    &=T(t)+E_y(y)+E_s(1),\\
    \boldsymbol{m}^{M}_{i}
    &={P_{\mathrm{cond}}}(\boldsymbol{c}^{M}_{i})
      +T(t+1)+E_\sigma(\sigma)+E_s(2),
\end{aligned}
\label{eq:conditions}
\end{equation}
{where $i$ indexes spatial tokens in the condition and target representations.} The offset $T(t+1)$ places the two stages in disjoint ranges of the timestep embedding. In each DiT block, the standard adaLN-Zero projection maps $\boldsymbol{m}^{L}$ or $\boldsymbol{m}^{M}_{i}$ to shift, scale, and residual-gating parameters. Stage~1 therefore shares the same modulation across all tokens, whereas Stage~2 produces token-specific modulation from $\boldsymbol{m}^{M}_{i}$, allowing each deep token to directly control the corresponding shallow token. {The stages share attention, FFNs, and AdaLN modulation MLPs. Stage-specific input and output projections accommodate the target dimensions, while separate condition mappings handle class labels and spatial representations.}

\setlength{\dbltextfloatsep}{12pt plus 2pt minus 2pt}
\begin{table*}[!t]
\centering
\small
\setlength{\tabcolsep}{3pt}
\renewcommand{\arraystretch}{0.94}
\benchmarktable{lrrrrrrrrrrr}{12}{%
\toprule
 & & & & \multicolumn{4}{c}{Without CFG guidance} & \multicolumn{4}{c}{With CFG or Autoguidance} \\
\cmidrule(lr){5-8}\cmidrule(l){9-12}
\textbf{Method} & \cellcolor{myblue!20}\textbf{Epochs} & \cellcolor{myblue!20}\textbf{Dec. Params} & \textbf{Params} & \textbf{FID}$\downarrow$ & \textbf{IS}$\uparrow$ & \textbf{Prec.}$\uparrow$ & \textbf{Rec.}$\uparrow$ & \textbf{FID}$\downarrow$ & \textbf{IS}$\uparrow$ & \textbf{Prec.}$\uparrow$ & \textbf{Rec.}$\uparrow$ \\

\midrule
\multicolumn{12}{@{}l}{\textit{Pixel diffusion}} \\
ADM~\citep{adm} & 400 & 0 & 554M & 10.94 & 101.0 & 0.69 & 0.63 & 3.94 & 215.8 & 0.83 & 0.53\\
SiD U-ViT/2~\citep{hoogeboom2023simple} & -- & 0 & 2B & 2.77 & -- & -- & -- & 2.44 & -- & -- & --\\
SiD2 U-ViT/2~\citep{sid2} & -- & 0 & -- & -- & -- & -- & -- & 1.73 & -- & -- & --\\
JiT-L~\citep{li2025jit} & 200 & 0 & 459M & 16.21 & -- & -- & -- & 2.79 & -- & -- & --\\
LF-DiT-L~\citep{baade2026latentforcing} & 200 & 0 & 465M & 7.20 & -- & -- & -- & 2.48 & -- & -- & --\\
RG-XL/16~\citep{pan2026rg} & 600 & 0 & 725M & -- & -- & -- & -- & 1.55 & 293.8 & 0.80 & 0.64\\
\addlinespace[3pt]
\multicolumn{12}{@{}l}{\textit{Latent diffusion}} \\
DiT-XL/2~\citep{peebles2023dit} & 1400 & 49M & 675M & 9.62 & 121.5 & 0.67 & 0.67 & 2.27 & 278.2 & 0.83 & 0.57\\
SiT-XL/2~\citep{ma2024sit} & 1400 & 49M & 675M & 8.61 & 131.7 & 0.68 & 0.67 & 2.06 & 270.3 & 0.82 & 0.59\\
REPA~\citep{yu2025repa} & 800 & 49M & 675M & 5.90 & 157.8 & 0.70 & 0.69 & 1.42 & 305.7 & 0.80 & 0.64\\
MAR + DeTok~\citep{yang2025detok} & 800 & 86M & 479M & 1.86 & 238.6 & 0.82 & 0.61 & 1.35 & 304.1 & 0.81 & 0.62\\
LightningDiT~\citep{yao2025vavae} & 800 & 41M & 675M & 2.17 & 205.6 & 0.77 & 0.65 & 1.35 & 295.3 & 0.79 & 0.65\\
REPA-E~\citep{repa-e} & 800 & 41M & 675M & 1.69 & 219.3 & 0.77 & 0.67 & 1.12 & 302.9 & 0.79 & 0.66\\
MAR + UniFlow~\citep{yue2025uniflow} & 400 & 300M & 479M & 2.45 & 228.0 & -- & -- & 1.85 & 290.0 & -- & --\\
LightningDiT + IG~\citep{ig} & 680 & 41M & 678M & 1.34 & 229.3 & 0.78 & 0.65 & 1.19 & 269.0 & 0.79 & 0.66\\
RAE~\citep{zheng2025rae} & 800 & 415M & 839M & 1.51 & 242.9 & 0.79 & 0.63 & 1.13 & 262.6 & 0.78 & 0.67\\
\midrule
LightningDiT~\citep{yao2025vavae} & 64 & 41M & 675M & 5.14 & 130.2 & 0.76 & 0.62 & -- & -- & -- & --\\
LightningDiT + IG~\citep{ig} & 60 & 41M & 678M & 2.42 & 173.7 & 0.79 & 0.62 & -- & -- & -- & --\\
RAE~\citep{zheng2025rae} & 20 & 415M & 839M & 3.71 & 198.7 & 0.86 & 0.50 & -- & -- & -- & --\\
\rowcolor{myblue!6} \textbf{HyC-DiT-XL + RevFlow (Ours)} & 5 & 89M & 690M & \textbf{1.96} & 237.1 & 0.71 & 0.61 & -- & -- & -- & --\\
\rowcolor{myblue!6} \textbf{HyC-DiT-H + RevFlow (Ours)} & 5 & 89M & 1.55B & \textbf{1.74} & 238.0 & 0.71 & 0.62 & -- & -- & -- & --\\
\addlinespace[4pt]
REPA~\citep{yu2025repa} & 80 & 49M & 675M & 7.90 & 122.6 & 0.70 & 0.65 & -- & -- & -- & --\\
REPA + E2E-VAE~\citep{repa-e} & 80 & 41M & 675M & 3.46 & 159.8 & 0.77 & 0.63 & 1.67 & 266.3 & 0.80 & 0.63\\
RAE~\citep{zheng2025rae} & 80 & 415M & 839M &  2.16 &  214.8 & 0.82  &  0.59 &  1.74 & 235.0  & 0.81  & 0.60 \\
\rowcolor{myblue!6} \textbf{HyC-DiT-XL + RevFlow (Ours)} & 80 & 89M & 690M & \textbf{1.36} & 257.6 & 0.71 & 0.64 & 1.33 & 254.0 & 0.70 & 0.65\\
\rowcolor{myblue!6} \textbf{HyC-DiT-H + RevFlow (Ours)} & 80 & 89M & 1.55B & \textbf{1.17} & 253.0 & 0.70 & 0.66 & 1.09 & 250.4 & 0.69 & 0.66\\
\bottomrule
}
\caption{\textbf{Class-conditional ImageNet 256$\times$256.} Lower blocks compare our 5-epoch models with 20--64-epoch baselines and all 80-epoch models together. \textcolor{black}{RevFlow-XL achieves better FID than RAE (1.96 vs.\ 2.16) with approximately 11$\times$ lower cumulative forward compute.} Dec.\ Params counts pixel-decoder parameters (0 for pixel-space models). RevFlow-XL/H share a Transformer backbone across generation stages and use REPA guidance without CFG; the right columns additionally use AutoGuidance~\citep{AG}.}
\label{tab:comparison_perf}
\end{table*}

\subsection{HyC-DiT Training: Clean-Target Prediction}
\label{sec:prediction_space}
\begin{figure}[t]
\centering
\includegraphics[width=0.94\linewidth]{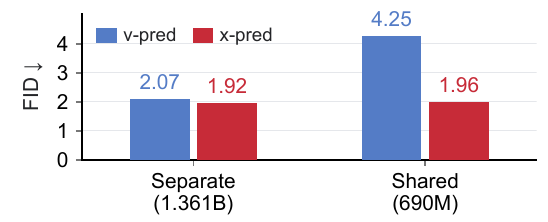}
\caption{\textcolor{black}{\textbf{Prediction target and parameter sharing.} Weight sharing degrades $v$-prediction, whereas clean-target prediction preserves generation quality with approximately half the parameters.}}
\label{fig:prediction_sharing}
\end{figure}

{Both stages use clean-target prediction~\citep{li2025jit} as defined in Eq.~\eqref{eq:stage_predictions}. For either representation target, the model could instead predict the corresponding velocity field. These parameterizations describe the same flow but differ in optimization under parameter sharing.} Clean-target prediction outputs denoised representations anchored near their conditional means $\mathbb{E}[\boldsymbol{z}^{\ell}|\boldsymbol{c}^{\ell}]$ in both stages, whereas velocity prediction couples stage-specific residuals with high-dimensional noise $\boldsymbol{\epsilon}^{\ell}$ (Appendix~\ref{sec:theory}), causing interference in a shared backbone. {Figure~\ref{fig:prediction_sharing} shows that clean-target prediction retains comparable quality with shared and separate networks (1.96 vs.\ 1.92 FID), while reducing parameters from 1.361B to 690M. Velocity prediction instead degrades from 2.07 to 4.25 FID under sharing, motivating our choice of clean-target prediction.}

\noindent\textbf{Training objective.}
{For the clean-target predictions in Eq.~\eqref{eq:stage_predictions}, the interpolation in Eq.~\eqref{eq:interpolation} yields the flow-matching loss}
$\mathcal{L}_{\mathrm{FM}}^{\ell}
=\mathbb{E}\|t^{-1}
(\hat{\boldsymbol{z}}^{\ell}-\boldsymbol{z}^{\ell})\|_2^2$. We jointly optimize the two stages as
\begin{equation}
    \mathcal{L}
    =
    w_c\,\mathcal{L}_{\mathrm{FM}}^{L}
    +
    \mathcal{L}_{\mathrm{FM}}^{M},
    \label{eq:total_loss}
\end{equation}
where $w_c$ balances the two stages. %
{Following RAEv2~\citep{singh2026raev2}, an auxiliary head at an intermediate Transformer block predicts the same clean target as the final head. It receives additional supervision during training} {and provides REPA guidance during sampling.}

\noindent\textbf{Sampling.}
Starting each stage from independent Gaussian noise, RevFlow first generates $\boldsymbol{z}^{L}$ conditioned on $y$, then generates $\boldsymbol{z}^{M}$ conditioned on the generated deep representation, and finally decodes $\boldsymbol{z}^{M}$ to pixels. %
{REPA guidance extrapolates from the auxiliary prediction toward the final prediction, using both outputs from the same backbone evaluation.} For multi-step ImageNet generation, we take 25 Euler steps per stage. With AutoGuidance, each step also evaluates a weak model.

\subsection{Native Representation Matching}
\label{sec:one_step}
{Representation-based matching compares distributions in pretrained feature spaces~\citep{yang2026fdloss,feng2026rdm,deng2026drifting}. For conventional VAE-based generators, this typically requires decoding sampled latents into images and extracting their features. RevFlow directly generates encoder representations, allowing native matching in the target representation spaces without this decode--encode process.} \textcolor{black}{We describe one-step-per-stage adaptation for two-stage RevFlow below and evaluate single-stage RAE in Sec.~\ref{sec:one_step_results}.}

Starting from independent Gaussian noise at $t=1$, the one-step cascade produces
\begin{equation}
\begin{aligned}
    \hat{\boldsymbol{z}}^{L}
    &=f_{\theta}(\boldsymbol{\epsilon}^{L},1,y), \qquad
    \hat{\boldsymbol{z}}^{M}
    &=f_{\theta}(\boldsymbol{\epsilon}^{M},1,\hat{\boldsymbol{z}}^{L}).
\end{aligned}
\label{eq:one_step_path}
\end{equation}
Reference representations are obtained from an exponential moving average (EMA) generator using noised real representations and are treated as stop-gradient targets. Within each class $y$, we use a drifting loss~\citep{deng2026drifting}, $\mathcal{L}_{\mathrm{stat}}^{y}$, to match multiscale channel-wise window statistics, and IMQ-MMD~\citep{gretton2012mmd}, $\mathcal{L}_{\mathrm{IMQ}}^{\ell,y}$, to align the marginal distribution at each level $\ell\in\{L,M\}$. Both operate directly on the generated representations without sample-wise pairing or an additional learned feature network. We retain clean-target flow matching as a regularizer and optimize
\begin{equation}
\mathcal{L}_{\mathrm{NM}}
=\mathbb{E}_{y}\big[
\lambda_{s}\mathcal{L}_{\mathrm{stat}}^{y}
+\textstyle\sum_{\ell}
\lambda_{\ell}\mathcal{L}_{\mathrm{IMQ}}^{\ell,y}
\big]
+\lambda_{\mathrm{FM}}{\mathcal{L}},
\label{eq:native_matching_loss}
\end{equation}
{where $\mathcal{L}$ is the flow-matching loss in Eq.~\eqref{eq:total_loss} and the $\lambda$ coefficients weight the terms.} After adaptation, RevFlow {needs one evaluation per stage plus the frozen decoder}.

\section{Experiments}
\label{sec:experiments}
\setlength{\intextsep}{8pt plus 2pt minus 2pt}
\setlength{\textfloatsep}{12pt plus 2pt minus 2pt}

We evaluate RevFlow on class-conditional ImageNet at $256\times256$, and on text-to-image generation at $512\times512${, covering convergence, generation quality, reconstruction, few-step sampling, and ablations}.

\subsection{Main results}
\label{sec:main_results}

For the five-epoch ImageNet runs, we train RevFlow-XL and RevFlow-H with Muon~\citep{jordan2024muon} and AdamW at a global batch size of 256. We compute FID on 50,000 generated images (50 per class) using ImageNet training-set reference statistics.

\noindent\textbf{ImageNet 256$\times$256.}
Table~\ref{tab:comparison_perf} shows that RevFlow-XL (690M) and RevFlow-H (1.55B) reach \textbf{1.96} and \textbf{1.74 FID} after only five epochs, using a unified generative model with a Transformer backbone shared across stages.
A training example costs 1,652.92 forward GFLOPs for RevFlow-XL and 1,134.59 for RAE. Over 5 and 80 epochs, respectively, RevFlow-XL uses about one eleventh of RAE's cumulative forward FLOPs and achieves a lower FID (1.96 vs.\ 2.16).
At 80 epochs, RevFlow-H achieves \textbf{1.17 FID} without CFG and \textbf{1.09} with AutoGuidance~\citep{AG}, outperforming RAE's 800-epoch results (1.51/1.13) with 10$\times$ fewer epochs.

Generating shallow representations enables RevFlow to combine strong generation quality with faithful reconstruction using a lightweight decoder. Decoding shallow representations extracted from ImageNet validation images yields 30.25 dB PSNR with 89M decoder parameters, compared with 18.09 dB and 415M for RAE (Fig.~\ref{fig:intro}). Pixel-space generation avoids autoencoder reconstruction error but can require longer training: JiT-L reaches 16.21 FID without CFG after 200 epochs.

\begin{table}[!htbp]
\centering
\footnotesize
\setlength{\tabcolsep}{2.2pt}
\renewcommand{\arraystretch}{1.12}
\benchmarktable{lrrrrrrrr}{9}{%
\toprule
Method & Ep. & Inc. & CNX & DINO & MAE & SigL & CLIP & $\mathrm{FDr}^{6}\!\downarrow$ \\
\midrule
SiT-XL/2~\citep{ma2024sit} & 800 & 1.26 & 2.02 & 7.89 & 5.62 & 16.14 & 17.69 & 8.44 \\
DDT-XL~\citep{ddt} & 800 & 0.75 & 1.02 & 4.26 & 4.11 & 10.16 & 13.86 & 5.70 \\
LightningDiT~\citep{yao2025vavae} & 800 & 0.85 & 1.09 & 3.76 & 3.02 & 8.47 & 10.21 & 4.57 \\
RAE-XL~\citep{zheng2025rae} & 800 & 0.69 & 1.79 & 2.11 & 3.30 & 3.79 & 7.87 & 3.26 \\
REPA-E~\citep{repa-e} & 800 & 0.70 & 1.28 & 2.44 & 2.52 & 5.04 & 6.28 & 3.04 \\
\midrule
RAE (DINOv2-B)$^{*}$ & 80 & 1.32 & 2.33 & 3.14 & 6.33 & 7.56 & -- & -- \\
RAE (DINOv3-B)$^{*}$ & 80 & 1.33 & 2.82 & 3.89 & 6.80 & 8.17 & -- & -- \\
DiT + E2E-VA-VAE$^{*}$ & 80 & 2.64 & 5.90 & 8.59 & 6.33 & 16.30 & -- & -- \\
DiT + FLUX.2-VAE$^{*}$ & 80 & 2.76 & 4.65 & 8.56 & 5.15 & 17.41 & -- & -- \\
\addlinespace[4pt]
\rowcolor{myblue!6} RevFlow-XL (Ours) & 5 & 1.15 & 1.28 & 2.91 & 6.42 & 8.73 & 6.63 & 4.52 \\
\rowcolor{myblue!6} RevFlow-XL (Ours) & 80 & 0.82 & 0.92 & 1.83 & 4.46 & 5.52 & 4.73 & 3.05 \\
\rowcolor{myblue!6} \textbf{RevFlow-H (Ours)} & 80 & \textbf{0.72} & \textbf{0.81} & \textbf{1.34} & \textbf{3.61} & \textbf{4.26} & \textbf{3.63} & \textbf{2.40} \\
\bottomrule
}
\captionsetup{font=small}
\caption{\textbf{Generation quality across representation spaces.} On ImageNet 256$\times$256, we report normalized Representation Fr\'echet Distance ($\mathrm{FD}_{r}$) in six feature spaces: Inception (Inc.), ConvNeXt (CNX), DINOv2 (DINO), MAE, SigLIP (SigL), and CLIP. $\mathrm{FDr}^{6}$ is their mean, following FD-loss~\citep{yang2026fdloss}. Ep.\ denotes training epochs. $^{*}$ from NanoGen~\citep{leng2026diffusionbench}.}
\label{tab:fdr6}
\end{table}

\noindent\textbf{Evaluation across representation spaces.}
FID evaluates only the Inception feature space. Table~\ref{tab:fdr6} additionally reports normalized Fr\'echet distances across six representation spaces and their mean, $\mathrm{FDr}^{6}$, following FD-loss~\citep{yang2026fdloss}. RevFlow-XL improves from 4.52 after 5 epochs to 3.05 after 80 epochs. RevFlow-H reaches \textbf{2.40}, the lowest mean among the listed methods, compared with 3.04 for REPA-E and 3.26 for RAE-XL after 800 epochs. Compared with REPA-E, RevFlow-H improves four of the six representation scores, {indicating gains beyond Inception features}.

\subsection{Few-step generation via native matching}
\label{sec:one_step_results}

\begin{table}[!htbp]
\centering
\small
\setlength{\tabcolsep}{2pt}
\renewcommand{\arraystretch}{1.05}
\benchmarktable{llcr}{4}{%
\toprule
Method & Space & NFE & FID$\downarrow$ \\
\midrule
\textcolor{gray}{MeanFlow~\citep{geng2025meanflow}} & \textcolor{gray}{VAE} & \textcolor{gray}{1} & \textcolor{gray}{3.43} \\
\textcolor{black}{LightningDiT (800ep)} & \textcolor{black}{VAE} & \textcolor{black}{1} & \textcolor{black}{264.56} \\
\textcolor{black}{\quad + NM} & \textcolor{black}{VAE} & \textcolor{black}{1} & \textcolor{black}{291.68} \\
\midrule
\textcolor{black}{RAE} & \textcolor{black}{Encoder} & \textcolor{black}{1} & \textcolor{black}{284.19} \\
\textcolor{black}{\quad + NM} & \textcolor{black}{Encoder} & \textcolor{black}{1} & \textcolor{black}{3.46} \\

\rowcolor{myblue!6} \textbf{Ours} & Encoder & 2 & 281.78\\
\rowcolor{myblue!6} \quad + NM & Encoder & 2 & \textbf{2.85} \\
\bottomrule
}
\captionsetup{font=small}
\caption{\textcolor{black}{\textbf{Native matching on ImageNet 256$\times$256.} NM enables few-step generation in both RAE and RevFlow. Each pair fixes NFE before and after adaptation. NFE excludes the decoder; adapted models use no CFG. MeanFlow is an external reference.}}
\label{tab:one_step}
\end{table}

{\color{black}
Table~\ref{tab:one_step} tests native representation matching (NM) on single-stage RAE and two-stage RevFlow, comparing each generator before and after adaptation at a fixed NFE.
RAE uses 1 NFE and RevFlow uses 2 (one per stage), excluding the frozen pixel decoder.
Adapted models use no CFG. The RevFlow-XL variant used for native matching has a deep-stage DDT head (852M generator parameters). We initialize it from an 80-epoch checkpoint and adapt it for 1,000 updates.

\noindent\textbf{Transfer across semantic generators.}
NM improves RAE from 284.19 to \textbf{3.46 FID} at 1 NFE and RevFlow-XL from 281.78 to \textbf{2.85 FID} at 2 NFE.
These gains across single-stage and two-stage generators demonstrate applicability beyond RevFlow.
Both match generated representations directly, without pixel decoding or re-encoding.

\noindent\textbf{Role of the representation space.}
Under the same recipe, directly matching the VAE latents of the 800-epoch LightningDiT checkpoint worsens FID from 264.56 to 291.68 at 1 NFE.
This contrast supports semantic representations as both generation targets and matching spaces; the tested VAE control does not benefit.
\par}

\subsection{Ablation studies}
\label{sec:ablations}

\newcommand{\pend}{--}
\begin{table}[!htbp]
\centering
\small
\setlength{\tabcolsep}{4pt}
\renewcommand{\arraystretch}{1.04}
\newcommand{\grp}[1]{\addlinespace[3pt]\multicolumn{4}{@{}l}{\footnotesize\textit{#1}}\\[-0.5pt]}
\benchmarktable{lrrr}{4}{%
\toprule
Variant & \cellcolor{myblue!20}Dec.\ Params & \cellcolor{myblue!20}PSNR$\uparrow$ & FID$\downarrow$ \\

\midrule
\rowcolor{myblue!6} \textbf{RevFlow}: $\boldsymbol{z}^{L}_{23}\!\to\!\boldsymbol{z}^{M}_{7}\!\to\!\boldsymbol{x}$ & \textcolor{gray}{89M} & \textcolor{gray}{30.25} & 1.96 \\
\grp{(a) Generation path}
\textcolor{black}{Deep only: $\boldsymbol{z}^{L}_{23}\!\to\!\boldsymbol{x}$} & 89M & 19.87 & 2.65 \\
Shallow only: $\boldsymbol{z}^{M}_{7}\!\to\!\boldsymbol{x}$ & \textcolor{gray}{89M} & \textcolor{gray}{30.25} & 15.94 \\
VAE target: $\boldsymbol{z}^{L}_{23}\!\to$ SD-VAE & 49M & 23.40 & 2.94 \\
Three stages: $+\,\boldsymbol{z}_{16}$ & \textcolor{gray}{89M} & \textcolor{gray}{30.25} & 2.51 \\
\grp{(b) Stage-2 condition injection}
+ Pooled context tokens & \textcolor{gray}{89M} & \textcolor{gray}{30.25} & 2.02 \\
Spatial injection: 1/7/14/21 & \textcolor{gray}{89M} & \textcolor{gray}{30.25} & 1.96 \\
Spatial injection: first layer & \textcolor{gray}{89M} & \textcolor{gray}{30.25} & 2.03 \\
\grp{(c) Condition noise}
$\sigma=0$ & \textcolor{gray}{89M} & \textcolor{gray}{30.25} & 2.10 \\
\grp{(d) Prediction target}
$v$-prediction & \textcolor{gray}{89M} & \textcolor{gray}{30.25} & \textcolor{black}{4.25} \\
\bottomrule
}
\captionsetup{font=small}
\caption{\textcolor{black}{\textbf{Ablations on ImageNet 256$\times$256.} All variants use DINOv3-L, DiT-XL, five-epoch training, and 50 NFE with windowed REPA guidance. RevFlow's deep-to-shallow path combines the lowest FID among generation paths with faithful reconstruction. Dec.\ Params counts pixel-decoder parameters; PSNR measures ImageNet validation reconstruction.}}
\label{tab:ablation_main}
\end{table}

\textcolor{black}{Table~\ref{tab:ablation_main} examines generation paths (a), condition injection (b), condition noise (c), and prediction targets (d).}

\noindent\textbf{Generation path.}
\textcolor{black}{Table~\ref{tab:ablation_main}(a) compares three paths with the same DINOv3-L encoder, DiT-XL configuration, clean-target prediction, and five-epoch training recipe. All use 89M pixel decoders and 50 total sampling steps under the common guidance protocol. Deep-only generation gives 2.65 FID and 19.87 dB reconstruction PSNR. Shallow-only generation retains 30.25 dB PSNR but yields 15.94 FID. RevFlow reduces FID to 1.96 with the same shallow target and decoder. RevFlow achieves both high-quality generation and faithful reconstruction, validating the effectiveness of its deep-to-shallow generation path. Replacing the shallow target with SD-VAE latents gives 2.94 FID; adding a third stage through layer 16 gives 2.51 FID. Both underperform the default path.}

\noindent\textbf{Condition injection.}
Table~\ref{tab:ablation_main}(b) evaluates additional pooled context tokens and the Transformer blocks at which spatial conditioning is applied. Adding pooled context tokens gives 2.02 FID, providing no improvement over 1.96. Restricting spatial injection to layers 1/7/14/21 gives 1.96 FID, matching the default; restricting it to the first layer gives 2.03.

\noindent\textbf{Condition noise.}
During training, Stage~2 receives deep representations extracted from real images; during sampling, it receives the deep representations generated by Stage~1. Removing condition noise increases FID from 1.96 to 2.10 (Table~\ref{tab:ablation_main}(c)), supporting its use in the current recipe.

\noindent\textbf{Target layers.}
\textcolor{black}{An additional shallow-depth study scans shallow targets in DINOv3-S with the deep target fixed at L12 and an 87.12M-parameter decoder.} Moving the shallow target from L1 to L3 and L5 improves FID from 4.36 to 3.14 and 2.69, while reconstruction PSNR decreases from 30.19 to 28.32 and 25.48 dB. This illustrates the trade-off between generation quality and reconstruction fidelity when selecting the shallow target.

\subsection{Text-to-image generation}
\label{sec:t2i_results}

\noindent\textbf{Setup and baseline.}
For text-to-image generation at $512\times512$, embeddings from a frozen Flan-T5 encoder~\citep{chung2024flan} replace the class condition in the first stage; the subsequent stage remains conditioned on the generated deep representation. Because published methods use different datasets, we reproduce RAE under our training setup and compare both methods at the same resolution from 5k to 35k iterations. Table~\ref{tab:t2i_benchmark} includes published latent- and pixel-space models for context{, with our reproduced RAE as the primary baseline}.

\begin{table}[!htbp]
\centering
\small
\renewcommand{\arraystretch}{1.03}
\benchmarktable{lrrr}{4}{%
\toprule
Method & Res. & GenEval$\uparrow$ & DPG-Bench$\uparrow$ \\
\midrule
SDXL~\citep{podell2023sdxl} & $1024^2$ & 0.55 & 74.7 \\
PixArt-$\Sigma$~\citep{chen2024pixartsigma} & $1024^2$ & 0.54 & 80.5 \\
FLUX.1-dev~\citep{bfl2024flux} & $1024^2$ & 0.67 & 84.0 \\
PixelFlow~\citep{pixelflow} & $512^2$ & 0.60 & 77.9 \\
PixNerd~\citep{pixnerd} & $512^2$ & 0.73 & 80.9 \\
\midrule
RAE (reproduced) & $512^2$ & 0.74 & 77.6 \\
\rowcolor{myblue!6} \textbf{RevFlow (Ours)} & $512^2$ & \textbf{0.78} & \textbf{81.6} \\
\bottomrule
}
\captionsetup{font=small}
\caption{\textbf{Text-to-image benchmarks.} Overall GenEval~\citep{ghosh2023geneval} and DPG-Bench~\citep{hu2024ella} scores. Res.: evaluation resolution.}
\label{tab:t2i_benchmark}
\end{table}

\begin{figure}[!htb]
\centering
\includegraphics[width=\linewidth]{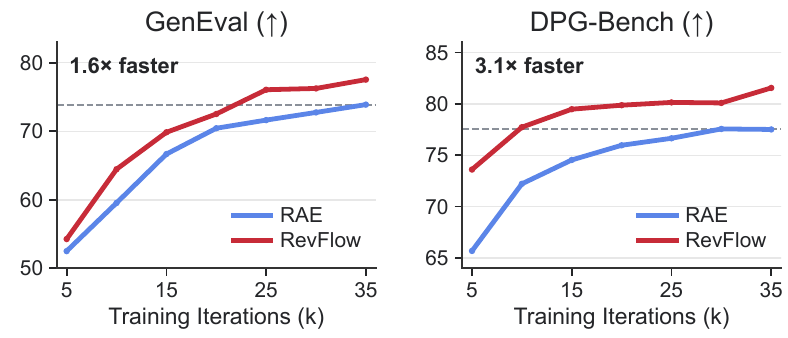}
\caption{\textbf{Text-to-image convergence.} GenEval is reported as a percentage. Markers show measured checkpoints. Dashed lines mark RAE's best recorded scores.}
\label{fig:t2i_convergence}
\end{figure}

\noindent\textbf{Results and convergence.}
RevFlow outperforms the reproduced RAE at every evaluated checkpoint on both benchmarks (Fig.~\ref{fig:t2i_convergence}). At 35k iterations, it reaches 77.54\% on GenEval and 81.55 on DPG-Bench, improving over RAE by 3.65 and 4.04 points. {Both scores still rise from 30k to 35k (from 76.25\% and 80.09), suggesting gains from longer training.}

\section{Conclusion}
\label{sec:conclusion}

RevFlow models the reverse of a frozen encoder hierarchy as successive generative transitions, improving convergence and reconstruction with a Transformer backbone shared across stages and a lightweight pixel decoder. Clean-target prediction enables effective parameter sharing, while native representation matching supports one-step generation per stage without pixel decoding and re-encoding during adaptation. These results establish encoder hierarchies as useful structure for organizing image generation.

\clearpage
\appendix
\section{Theoretical Analysis of Reverse Encoding}
\label{sec:theory}

This appendix examines how the deep representation guides shallow-feature generation in Sec.~\ref{sec:path}. We first separate the conditional mean, which depends on the deep features alone, from the residual that must be denoised. Proposition~\ref{prop:flow_variance} then shows how the posterior covariance of this residual governs the sensitivity of the optimal predictor to its noisy input. Finally, Sec.~\ref{sec:theory_evidence} verifies these properties empirically on DINOv3-L features and connects them to our ablations.

\paragraph{Setup and definitions.}
\revadd{Along the frozen encoder $\boldsymbol{x}=\boldsymbol{h}^{0}\to\dots\to\boldsymbol{h}^{M}\to\dots\to\boldsymbol{h}^{L}$ ($M<L$), deterministic blocks suppress detail while retaining semantics. RevFlow reverses this path: Stage~1 generates $\boldsymbol{z}^{L}=\boldsymbol{h}^{L}$ from $y$, Stage~2 generates $\boldsymbol{z}^{M}=P_{\mathrm{target}}(\boldsymbol{h}^{M})$ from $\boldsymbol{z}^{L}$, and a decoder maps $\boldsymbol{z}^{M}$ to $\boldsymbol{x}$. We write $\boldsymbol{a}\in\mathbb{R}^{d}$ ($\mathbb{E}\|\boldsymbol{a}\|^{2}<\infty$) for a clean stage target and $\boldsymbol{c}=\boldsymbol{h}^{L}$ for the deep feature.}

\begin{definition}[\revadd{Mean guidance and clean residual}]
\label{def:residual}
For a target $\boldsymbol{a}\in\mathbb{R}^{d}$ and condition $\boldsymbol{u}$, let $V(\boldsymbol{a})=\mathbb{E}\|\boldsymbol{a}-\mathbb{E}\boldsymbol{a}\|^{2}$ and $V(\boldsymbol{a}|\boldsymbol{u})=\mathbb{E}\|\boldsymbol{a}-\mathbb{E}[\boldsymbol{a}|\boldsymbol{u}]\|^{2}$. \revadd{Write $\boldsymbol{\mu}(\boldsymbol{u})=\mathbb{E}[\boldsymbol{a}|\boldsymbol{u}]$ for the conditional mean and $\boldsymbol{r}=\boldsymbol{a}-\boldsymbol{\mu}(\boldsymbol{u})$ for the zero-mean clean residual, so $\boldsymbol{a}=\boldsymbol{\mu}(\boldsymbol{u})+\boldsymbol{r}$ and $V(\boldsymbol{a}|\boldsymbol{u})=\mathbb{E}\|\boldsymbol{r}\|^{2}$. For the forward interpolant $\boldsymbol{a}_{t}=(1-t)\boldsymbol{a}+t\boldsymbol{\epsilon}$ ($t\in(0,1)$, $\boldsymbol{\epsilon}\sim\mathcal{N}(\boldsymbol{0},\mathbf{I}_{d})$ independent of $(\boldsymbol{a},\boldsymbol{u})$), subtracting $(1-t)\boldsymbol{\mu}(\boldsymbol{u})$ gives the residual interpolant $\boldsymbol{r}_{t}=\boldsymbol{a}_{t}-(1-t)\boldsymbol{\mu}(\boldsymbol{u})=(1-t)\boldsymbol{r}+t\boldsymbol{\epsilon}$. Let $\phi_{t}(\boldsymbol{a}_{t}|\boldsymbol{r})=\mathcal{N}(\boldsymbol{r}_{t};(1-t)\boldsymbol{r},t^{2}\mathbf{I}_{d})$ be the forward Gaussian kernel, $p_{t}(\boldsymbol{a}_{t}|\boldsymbol{u})=\int\phi_{t}(\boldsymbol{a}_{t}|\boldsymbol{r})\,P(d\boldsymbol{r}|\boldsymbol{u})$ the marginal density with score $\boldsymbol{s}_{t}(\boldsymbol{a}_{t}|\boldsymbol{u})=\nabla_{\boldsymbol{a}_{t}}\log p_{t}(\boldsymbol{a}_{t}|\boldsymbol{u})$, $\boldsymbol{a}^{*}(\boldsymbol{a}_{t},t,\boldsymbol{u})=\mathbb{E}[\boldsymbol{a}|\boldsymbol{a}_{t},\boldsymbol{u}]$ the optimal clean-target predictor (with Jacobian $(\nabla_{\boldsymbol{a}_{t}}\boldsymbol{a}^{*})_{ij}=\partial a^{*}_{i}/\partial a_{t,j}$), $\operatorname{Cov}(\boldsymbol{r}|\boldsymbol{a}_{t},\boldsymbol{u})=\mathbb{E}[\boldsymbol{r}\boldsymbol{r}^{\top}|\boldsymbol{a}_{t},\boldsymbol{u}]-\mathbb{E}[\boldsymbol{r}|\boldsymbol{a}_{t},\boldsymbol{u}]\mathbb{E}[\boldsymbol{r}|\boldsymbol{a}_{t},\boldsymbol{u}]^{\top}$ the posterior residual covariance, and $V(\boldsymbol{a}|\boldsymbol{a}_{t},\boldsymbol{u})=\mathbb{E}\|\boldsymbol{a}-\boldsymbol{a}^{*}\|^{2}=\mathbb{E}[\operatorname{tr}\operatorname{Cov}(\boldsymbol{r}|\boldsymbol{a}_{t},\boldsymbol{u})]$.}
\end{definition}
For any fixed predictor $g$, $V(\boldsymbol{a}|\boldsymbol{u})\le(1-R^{2})V(\boldsymbol{a})$, where $R^{2}=1-\mathbb{E}\|\boldsymbol{a}-g(\boldsymbol{u})\|^{2}/V(\boldsymbol{a})$, so measured $R^{2}$ values give upper bounds on $V$.

{\color{black}
\begin{proposition}[Mean--residual separation and denoiser gradient]
\label{prop:flow_variance}
Under Definition~\ref{def:residual}, for every $t\in(0,1)$, the optimal clean-target predictor satisfies
\begin{equation}
\begin{aligned}
\boldsymbol{a}^{*}(\boldsymbol{a}_{t},t,\boldsymbol{u})
&= \boldsymbol{\mu}(\boldsymbol{u})+\mathbb{E}[\boldsymbol{r}|\boldsymbol{a}_{t},\boldsymbol{u}]\\[-2pt]
&= \boldsymbol{\mu}(\boldsymbol{u})+\frac{\boldsymbol{r}_{t}+t^{2}\boldsymbol{s}_{t}(\boldsymbol{a}_{t}|\boldsymbol{u})}{1-t},
\end{aligned}
\label{eq:mean_residual_split}
\end{equation}
and its input Jacobian and expected trace are governed by the posterior residual variance:
\begin{equation}
\begin{aligned}
\nabla_{\boldsymbol{a}_{t}}\boldsymbol{a}^{*}(\boldsymbol{a}_{t},t,\boldsymbol{u})
&= \frac{1-t}{t^{2}}\operatorname{Cov}(\boldsymbol{r}|\boldsymbol{a}_{t},\boldsymbol{u}),\\[-2pt]
\mathbb{E}\big[\operatorname{tr}\nabla_{\boldsymbol{a}_{t}}\boldsymbol{a}^{*}(\boldsymbol{a}_{t},t,\boldsymbol{u})\big]
&= \frac{1-t}{t^{2}}V(\boldsymbol{a}|\boldsymbol{a}_{t},\boldsymbol{u}).
\end{aligned}
\label{eq:tweedie_jacobian}
\end{equation}
\end{proposition}
\begin{proof}
To obtain Eq.~\eqref{eq:mean_residual_split}, substitute $\boldsymbol{a}=\boldsymbol{\mu}(\boldsymbol{u})+\boldsymbol{r}$ into $\boldsymbol{a}^{*}=\mathbb{E}[\boldsymbol{a}|\boldsymbol{a}_{t},\boldsymbol{u}]$. Since $\mathbb{E}[\boldsymbol{\mu}(\boldsymbol{u})|\boldsymbol{a}_{t},\boldsymbol{u}]=\boldsymbol{\mu}(\boldsymbol{u})$, linearity gives the first equality:
\begin{equation*}
\boldsymbol{a}^{*}(\boldsymbol{a}_{t},t,\boldsymbol{u})
=\boldsymbol{\mu}(\boldsymbol{u})+\mathbb{E}[\boldsymbol{r}|\boldsymbol{a}_{t},\boldsymbol{u}].
\end{equation*}
Next, take the conditional expectation of the forward relation $\boldsymbol{r}_{t}=(1-t)\boldsymbol{r}+t\boldsymbol{\epsilon}$ given $(\boldsymbol{a}_{t},\boldsymbol{u})$ and write $\boldsymbol{\epsilon}^{*}=\mathbb{E}[\boldsymbol{\epsilon}|\boldsymbol{a}_{t},\boldsymbol{u}]$:
\begin{equation*}
\boldsymbol{r}_{t}=(1-t)\mathbb{E}[\boldsymbol{r}|\boldsymbol{a}_{t},\boldsymbol{u}]+t\boldsymbol{\epsilon}^{*}
\implies
\mathbb{E}[\boldsymbol{r}|\boldsymbol{a}_{t},\boldsymbol{u}]=\frac{\boldsymbol{r}_{t}-t\boldsymbol{\epsilon}^{*}}{1-t}.
\end{equation*}
Because the forward noise scale is $\sigma_{t}=t$, the score--noise identity $\boldsymbol{s}_{t}(\boldsymbol{a}_{t}|\boldsymbol{u})=-\boldsymbol{\epsilon}^{*}/\sigma_{t}=-\boldsymbol{\epsilon}^{*}/t$~\citep{song2021score} gives $-t\boldsymbol{\epsilon}^{*}=t^{2}\boldsymbol{s}_{t}(\boldsymbol{a}_{t}|\boldsymbol{u})$. Substituting this into the numerator yields the second equality in Eq.~\eqref{eq:mean_residual_split}.

To obtain Eq.~\eqref{eq:tweedie_jacobian}, note that $\boldsymbol{\mu}(\boldsymbol{u})$ is independent of $\boldsymbol{a}_{t}$, so $\nabla_{\boldsymbol{a}_{t}}\boldsymbol{a}^{*}=\nabla_{\boldsymbol{a}_{t}}\mathbb{E}[\boldsymbol{r}|\boldsymbol{a}_{t},\boldsymbol{u}]$. By Bayes' rule, write $\mathbb{E}[\boldsymbol{r}|\boldsymbol{a}_{t},\boldsymbol{u}]=\boldsymbol{N}(\boldsymbol{a}_{t})/p_{t}(\boldsymbol{a}_{t}|\boldsymbol{u})$ with $\boldsymbol{N}(\boldsymbol{a}_{t})=\int\boldsymbol{r}\phi_{t}(\boldsymbol{a}_{t}|\boldsymbol{r})\,P(d\boldsymbol{r}|\boldsymbol{u})$. Since $\nabla_{\boldsymbol{a}_{t}}\boldsymbol{r}_{t}=\mathbf{I}_{d}$, differentiating $\phi_{t}(\boldsymbol{a}_{t}|\boldsymbol{r})=(2\pi t^{2})^{-d/2}\exp(-\|\boldsymbol{r}_{t}-(1-t)\boldsymbol{r}\|^{2}/(2t^{2}))$ gives
\begin{equation*}
\nabla_{\boldsymbol{a}_{t}}\phi_{t}(\boldsymbol{a}_{t}|\boldsymbol{r})
=\frac{(1-t)\boldsymbol{r}-\boldsymbol{r}_{t}}{t^{2}}\,\phi_{t}(\boldsymbol{a}_{t}|\boldsymbol{r}).
\end{equation*}
Differentiating $p_{t}$ and $\boldsymbol{N}$ under the integral sign and dividing by $p_{t}(\boldsymbol{a}_{t}|\boldsymbol{u})$ re-expresses both ratios as posterior expectations:
\begin{equation*}
\begin{aligned}
\frac{\nabla_{\boldsymbol{a}_{t}}p_{t}}{p_{t}}
&= \frac{1-t}{t^{2}}\mathbb{E}[\boldsymbol{r}|\boldsymbol{a}_{t},\boldsymbol{u}]-\frac{\boldsymbol{r}_{t}}{t^{2}},\\[-2pt]
\frac{\nabla_{\boldsymbol{a}_{t}}\boldsymbol{N}}{p_{t}}
&= \frac{1-t}{t^{2}}\mathbb{E}[\boldsymbol{r}\boldsymbol{r}^{\top}|\boldsymbol{a}_{t},\boldsymbol{u}]
   -\frac{1}{t^{2}}\mathbb{E}[\boldsymbol{r}|\boldsymbol{a}_{t},\boldsymbol{u}]\boldsymbol{r}_{t}^{\top}.
\end{aligned}
\end{equation*}
Applying the quotient rule $\nabla_{\boldsymbol{a}_{t}}(\boldsymbol{N}/p_{t})=(\nabla_{\boldsymbol{a}_{t}}\boldsymbol{N})/p_{t}-(\boldsymbol{N}/p_{t})(\nabla_{\boldsymbol{a}_{t}}p_{t}/p_{t})^{\top}$, the two $-\frac{1}{t^{2}}\mathbb{E}[\boldsymbol{r}|\boldsymbol{a}_{t},\boldsymbol{u}]\boldsymbol{r}_{t}^{\top}$ terms cancel out:
\begin{equation*}
\begin{aligned}
\nabla_{\boldsymbol{a}_{t}}\boldsymbol{a}^{*}
&= \frac{1-t}{t^{2}}\Big(\mathbb{E}[\boldsymbol{r}\boldsymbol{r}^{\top}|\boldsymbol{a}_{t},\boldsymbol{u}]
   -\mathbb{E}[\boldsymbol{r}|\boldsymbol{a}_{t},\boldsymbol{u}]\mathbb{E}[\boldsymbol{r}|\boldsymbol{a}_{t},\boldsymbol{u}]^{\top}\Big)\\[-2pt]
&= \frac{1-t}{t^{2}}\operatorname{Cov}(\boldsymbol{r}|\boldsymbol{a}_{t},\boldsymbol{u}).
\end{aligned}
\end{equation*}
Taking the trace and expectation over $(\boldsymbol{a}_{t},\boldsymbol{u})$ gives the second line of Eq.~\eqref{eq:tweedie_jacobian}.
\end{proof}
}

\revadd{Because targets are encoded from \emph{clean} images, $P(\boldsymbol{r}|\boldsymbol{u})$ lies on sharp noise-free Dirac points. By Proposition~\ref{prop:flow_variance}, when $\boldsymbol{\mu}(\boldsymbol{u})$ is uninformative and $V(\boldsymbol{a}|\boldsymbol{u})$ is wide (direct $y\to\boldsymbol{h}^{M}$), clean modes overlap at high $t$ so residuals cancel ($\mathbb{E}[\boldsymbol{r}|\boldsymbol{a}_{t},y]\approx\boldsymbol{0}$, averaging out detail), while as $t\to 0$, $\operatorname{Cov}(\boldsymbol{r}|\boldsymbol{a}_{t},y)$ remains $\Theta(V(\boldsymbol{a}|y))$ at Dirac boundaries so $\|\nabla_{\boldsymbol{a}_{t}}\boldsymbol{a}^{*}\|_{\mathrm{op}}\propto\frac{1-t}{t^{2}}V(\boldsymbol{a}|y)\to\infty$, forcing stiff Dirac jumps rather than smooth score refinement. Conversely, when $\boldsymbol{\mu}(\boldsymbol{u})$ is informative and $V(\boldsymbol{a}|\boldsymbol{u})$ is narrow (local perturbation $\sigma_{r}^{2}\mathbf{I}_{d}$), $\operatorname{Cov}(\boldsymbol{r}|\boldsymbol{a}_{t},\boldsymbol{u})=O(t^{2})$ cancels $t^{-2}$ so $\|\nabla_{\boldsymbol{a}_{t}}\boldsymbol{a}^{*}\|_{\mathrm{op}}<1$, keeping the target mapping smooth around $\boldsymbol{\mu}(\boldsymbol{u})$~\citep{li2025jit} (also aiding backbone sharing in Fig.~\ref{fig:prediction_sharing} and one-step matching in Sec.~\ref{sec:one_step}). Reversing the path therefore keeps training smooth under two measurable encoder properties (Sec.~\ref{sec:theory_evidence}):}
\begin{itemize}[leftmargin=1.5em,itemsep=2pt,topsep=2pt]
\item[(C1)] \textbf{\revadd{Strong cross-level mean guidance:}} $V(\boldsymbol{a}|\boldsymbol{c})$ is small compared with $V(\boldsymbol{a}|y)$, so \revadd{$\boldsymbol{\mu}(\boldsymbol{c})$ anchors spatial structure and Stage~2 refines a narrow residual smoothly}.
\item[(C2)] \textbf{\revadd{Low-variance, class-organized deep target:}} $y$ explains a larger fraction of the variance of $\boldsymbol{c}$ than of $\boldsymbol{a}$, so \revadd{predicting $\boldsymbol{c}$ first operates at lower normalized residual variance $V(\boldsymbol{c}|y)/V(\boldsymbol{c})$}. When $y$ is also largely recoverable from $\boldsymbol{c}$, $V(\boldsymbol{a}|\boldsymbol{c},y)\approx V(\boldsymbol{a}|\boldsymbol{c})$, which justifies conditioning Stage~2 on $\boldsymbol{c}$ alone.
\end{itemize}

\revadd{Finally, since pixel variance $V(\boldsymbol{x}|\boldsymbol{h}^{\ell})$ grows with depth $\ell$ and lower-bounds any deterministic decoder ($\mathbb{E}\|\boldsymbol{x}-D(\boldsymbol{h}^{\ell})\|^{2}\ge V(\boldsymbol{x}|\boldsymbol{h}^{\ell})$), a} deterministic decoder from deep features must infer missing detail using its learned prior; the result may be averaged or biased toward patterns favored by the model rather than those in the original image~\citep{blau2018perception,mahendran2015understanding,dosovitskiy2016inverting}. This motivates a shallow final level. A shallower $\boldsymbol{a}$, however, weakens (C1), so the shallow level trades reconstruction against generation, and each additional level \revadd{with imperfect upstream prediction $\hat{\boldsymbol{c}}$ adds mean-guidance drift $\mathbb{E}\|\boldsymbol{\mu}(\boldsymbol{c})-\boldsymbol{\mu}(\hat{\boldsymbol{c}})\|^{2}$ (mitigated by condition noise $\sigma>0$ in Eq.~\eqref{eq:condition_noise}, and avoided within redundant similarity regimes in Fig.~\ref{fig:layer_similarity})}.

\subsection{Empirical Evidence from Encoder Features}
\label{sec:theory_evidence}

\begin{figure*}[t]
\centering
\includegraphics[width=\linewidth]{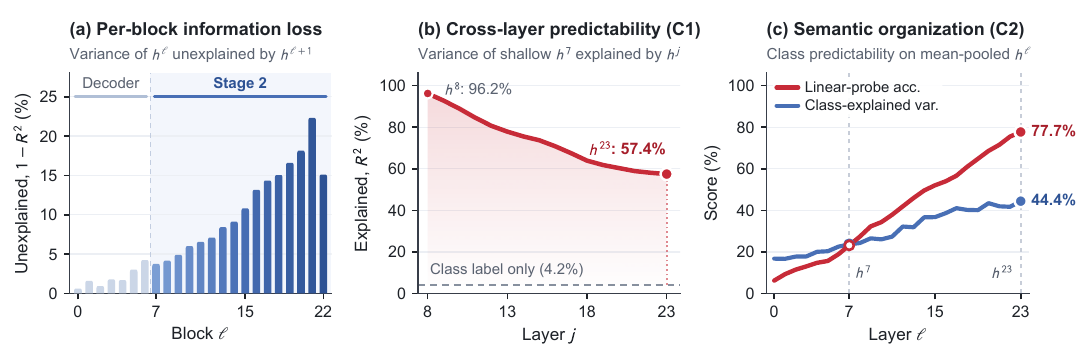}
\caption{\textbf{Layer statistics of DINOv3-L on ImageNet validation} (50K images; zero-based layers). $R^{2}$ is the fraction of a feature's variance explained by a linear prediction. \textbf{(a)} Variance of each layer that the next layer cannot explain ($1-R^{2}$), a proxy for what the block loses. Stage~2 restores what blocks 7--22 lose, and the decoder the rest. \textbf{(b)} Variance of $\boldsymbol{h}^{7}$ explained by layer $j$ or by the class label. \textbf{(c)} Linear-probe class accuracy, and the variance of mean-pooled features explained by the class.}
\label{fig:theory_measure}
\end{figure*}

\paragraph{Measurement protocol.}
We extract all 24 DINOv3-L layers for the 50K ImageNet validation images at $256\times256$ and sample 64 patch tokens per image. For each pair of layers, we fit a linear map from one layer's token to the other's at the same position and report its $R^{2}$. By Definition~\ref{def:residual}, $1-R^{2}$ upper-bounds the normalized residual variance $V/V(\boldsymbol{a})$; nonlinear predictors could only tighten it. The class predictor is the per-class mean. For Fig.~\ref{fig:theory_measure}c, we train an LDA classifier on mean-pooled features of 40 images per class and test on the remaining 10.

\paragraph{Condition (C1).}
For $\boldsymbol{a}=\boldsymbol{h}^{7}$ and $\boldsymbol{c}=\boldsymbol{h}^{23}$, the linear $R^{2}$ is 57.4\%, whereas the class alone explains 4.2\% (Fig.~\ref{fig:theory_measure}b). Hence $V(\boldsymbol{a}|\boldsymbol{c})\le0.426\,V(\boldsymbol{a})$, while $V(\boldsymbol{a}|y)=0.958\,V(\boldsymbol{a})$, the class mean being the optimal predictor from $y$. Stage~2 thus \revadd{receives strong spatial mean guidance and} generates less than half of the variation that direct generation from the class must produce. The $R^{2}$ decreases monotonically from 96.2\% for $\boldsymbol{c}=\boldsymbol{h}^{8}$ to 57.4\% for $\boldsymbol{h}^{23}$, and the variance lost per block grows from 0.6\% in the first block to 15--22\% in the last ones (Fig.~\ref{fig:theory_measure}a): (C1) weakens as the gap between levels widens.

\paragraph{Condition (C2).}
The class explains 18.6\% of the token variance of $\boldsymbol{h}^{23}$ but 4.2\% of that of $\boldsymbol{h}^{7}$. For mean-pooled features, the class-explained variance is 44.4\% versus 23.6\%, and linear-probe accuracy is 77.7\% versus 23.2\% (Fig.~\ref{fig:theory_measure}c). The deep level is thus organized by class \revadd{with substantially smaller normalized residual variance}, and the class is largely recoverable from it.

\paragraph{Ablations.}
The analysis predicts the outcome of each ablation under the same five-epoch recipe (Table~\ref{tab:ablation_main}).
\begin{itemize}[leftmargin=1.5em,itemsep=2pt,topsep=2pt]
\item \textbf{Removing the deep level} (\revadd{Prop.~\ref{prop:flow_variance}}). Generating the same shallow target directly from the class leaves all of $V(\boldsymbol{a}|y)$ to a single stage \revadd{under weak mean guidance ($R^{2}=4.2\%$)}. FID degrades from 1.96 to 15.94 with the target and decoder unchanged (Table~\ref{tab:ablation_main}a).
\item \textbf{Moving the shallow level} (C1). With the same decoder architecture, moving the DINOv3-S shallow target from layer 1 to 3 and 5 narrows the gap to the deep level and raises the reconstruction floor: FID improves from 4.36 to 3.14 and 2.69, while PSNR drops from 30.19 to 28.32 and 25.48~dB.
\item \textbf{Decoding the deep level}. \revadd{On DINOv3-L with a matched 89M decoder, decoding $\boldsymbol{z}^{L}_{23}$ directly ($\boldsymbol{z}^{L}_{23}\!\to\!\boldsymbol{x}$) drops PSNR from 30.25 to 19.87~dB and worsens FID from 1.96 to 2.65 (Table~\ref{tab:ablation_main}a). Similarly, the} final-layer RAE latent of DINOv2-B gives 18.09~dB PSNR with a 415M decoder and changes the hat color in Fig.~\ref{fig:intro}.
\item \textbf{Adding a level \revadd{and condition noise}.} A third stage at layer 16 \revadd{(within the late regime)} raises FID from 1.96 to 2.51 (Table~\ref{tab:ablation_main}a); here the accumulated \revadd{guidance drift} of the extra factor outweighs the shorter gaps. \revadd{Removing condition noise ($\sigma=0$) similarly degrades FID from 1.96 to 2.10 (Table~\ref{tab:ablation_main}c).}
\item \textbf{Replacing the shallow level with a VAE latent.} Generating SD-VAE latents from the same deep level gives 2.94 FID (Table~\ref{tab:ablation_main}a). Unlike $\boldsymbol{h}^{7}$, VAE latents are not part of the encoder path that produces $\boldsymbol{h}^{23}$.
\end{itemize}
Beyond final quality, the decomposition also trains fast: RevFlow reaches 1.96 FID after five epochs (Table~\ref{tab:comparison_perf}), and on text-to-image generation reaches the best GenEval and DPG-Bench scores of our RAE reproduction $1.6\times$ and $3.1\times$ faster (Fig.~\ref{fig:t2i_convergence}). The optimization recipe also contributes to this speed.


{
  \small
  \begin{thebibliography}{67}
\providecommand{\natexlab}[1]{#1}
\providecommand{\url}[1]{\texttt{#1}}
\expandafter\ifx\csname urlstyle\endcsname\relax
  \providecommand{\doi}[1]{doi: #1}\else
  \providecommand{\doi}{doi: \begingroup \urlstyle{rm}\Url}\fi

\bibitem[Achille and Soatto(2018)]{achille2018invariance}
Alessandro Achille and Stefano Soatto.
\newblock Emergence of invariance and disentanglement in deep representations.
\newblock \emph{Journal of Machine Learning Research}, 19\penalty0
  (50):\penalty0 1--34, 2018.

\bibitem[Albergo and Vanden-Eijnden(2022)]{albergo2023stochastic}
Michael~S Albergo and Eric Vanden-Eijnden.
\newblock Building normalizing flows with stochastic interpolants.
\newblock \emph{arXiv preprint arXiv:2209.15571}, 2022.

\bibitem[Baade et~al.(2026)Baade, Chan, Sargent, Chen, Johnson, Adeli, and
  Fei-Fei]{baade2026latentforcing}
Alan Baade, Eric~Ryan Chan, Kyle Sargent, Changan Chen, Justin Johnson, Ehsan
  Adeli, and Li Fei-Fei.
\newblock Latent forcing: Reordering the diffusion trajectory for pixel-space
  image generation.
\newblock \emph{arXiv preprint arXiv:2602.11401}, 2026.

\bibitem[Bansal et~al.(2023)Bansal, Borgnia, Chu, Li, Kazemi, Huang, Goldblum,
  Geiping, and Goldstein]{bansal2023cold}
Arpit Bansal, Eitan Borgnia, Hong-Min Chu, Jie~S. Li, Hamid Kazemi, Furong
  Huang, Micah Goldblum, Jonas Geiping, and Tom Goldstein.
\newblock Cold diffusion: Inverting arbitrary image transforms without noise.
\newblock In \emph{Advances in Neural Information Processing Systems}, 2023.

\bibitem[{Black Forest Labs}(2024)]{bfl2024flux}
{Black Forest Labs}.
\newblock {FLUX.1} [dev].
\newblock \url{https://huggingface.co/black-forest-labs/FLUX.1-dev}, 2024.

\bibitem[Blau and Michaeli(2018)]{blau2018perception}
Yochai Blau and Tomer Michaeli.
\newblock The perception-distortion tradeoff.
\newblock In \emph{Proceedings of the IEEE Conference on Computer Vision and
  Pattern Recognition}, pages 6228--6237, 2018.

\bibitem[Caron et~al.(2021)Caron, Touvron, Misra, J{\'e}gou, Mairal,
  Bojanowski, and Joulin]{caron2021dino}
Mathilde Caron, Hugo Touvron, Ishan Misra, Herv{\'e} J{\'e}gou, Julien Mairal,
  Piotr Bojanowski, and Armand Joulin.
\newblock Emerging properties in self-supervised vision transformers.
\newblock In \emph{Proceedings of the IEEE/CVF international conference on
  computer vision}, pages 9650--9660, 2021.

\bibitem[Chen et~al.(2024)Chen, Ge, Xie, Wu, Yao, Ren, Wang, Luo, Lu, and
  Li]{chen2024pixartsigma}
Junsong Chen, Chongjian Ge, Enze Xie, Yue Wu, Lewei Yao, Xiaozhe Ren, Zhongdao
  Wang, Ping Luo, Huchuan Lu, and Zhenguo Li.
\newblock {PixArt-$\Sigma$}: Weak-to-strong training of diffusion transformer
  for 4k text-to-image generation.
\newblock In \emph{European Conference on Computer Vision}, 2024.

\bibitem[Chen et~al.(2025)Chen, Ge, Zhang, Sun, and Luo]{pixelflow}
Shoufa Chen, Chongjian Ge, Shilong Zhang, Peize Sun, and Ping Luo.
\newblock Pixelflow: Pixel-space generative models with flow.
\newblock \emph{arXiv preprint arXiv:2504.07963}, 2025.

\bibitem[Chung et~al.(2024)Chung, Hou, Longpre, Zoph, Tay, Fedus, Li, Wang,
  Dehghani, Brahma, Webson, Gu, Dai, Suzgun, Chen, Chowdhery, Castro-Ros,
  Pellat, Robinson, Valter, Narang, Mishra, Yu, Zhao, Huang, Dai, Yu, Petrov,
  Chi, Dean, Devlin, Roberts, Zhou, Le, and Wei]{chung2024flan}
Hyung~Won Chung, Le Hou, Shayne Longpre, Barret Zoph, Yi Tay, William Fedus,
  Yunxuan Li, Xuezhi Wang, Mostafa Dehghani, Siddhartha Brahma, Albert Webson,
  Shixiang~Shane Gu, Zhuyun Dai, Mirac Suzgun, Xinyun Chen, Aakanksha
  Chowdhery, Alex Castro-Ros, Marie Pellat, Kevin Robinson, Dasha Valter,
  Sharan Narang, Gaurav Mishra, Adams Yu, Vincent Zhao, Yanping Huang, Andrew
  Dai, Hongkun Yu, Slav Petrov, Ed~H. Chi, Jeff Dean, Jacob Devlin, Adam
  Roberts, Denny Zhou, Quoc~V. Le, and Jason Wei.
\newblock Scaling instruction-finetuned language models.
\newblock \emph{Journal of Machine Learning Research}, 25\penalty0
  (70):\penalty0 1--53, 2024.

\bibitem[Deng et~al.(2026)Deng, Li, Li, Du, and He]{deng2026drifting}
Mingyang Deng, He Li, Tianhong Li, Yilun Du, and Kaiming He.
\newblock Generative modeling via drifting.
\newblock \emph{arXiv preprint arXiv:2602.04770}, 2026.

\bibitem[Dhariwal and Nichol(2021)]{adm}
Prafulla Dhariwal and Alexander Nichol.
\newblock Diffusion models beat gans on image synthesis.
\newblock \emph{Advances in neural information processing systems},
  34:\penalty0 8780--8794, 2021.

\bibitem[Dosovitskiy and Brox(2016)]{dosovitskiy2016inverting}
Alexey Dosovitskiy and Thomas Brox.
\newblock Inverting visual representations with convolutional networks.
\newblock In \emph{Proceedings of the IEEE Conference on Computer Vision and
  Pattern Recognition}, pages 4829--4837, 2016.

\bibitem[Esser et~al.(2021)Esser, Rombach, and Ommer]{esser2021taming}
Patrick Esser, Robin Rombach, and Bj{\"o}rn Ommer.
\newblock Taming transformers for high-resolution image synthesis.
\newblock In \emph{Proceedings of the IEEE/CVF Conference on Computer Vision
  and Pattern Recognition}, pages 12873--12883, 2021.

\bibitem[Feng et~al.(2026)Feng, Li, Zablocki, Cord, and Alahi]{feng2026rdm}
Lan Feng, Wuyang Li, {\'{E}}loi Zablocki, Matthieu Cord, and Alexandre Alahi.
\newblock Representation distribution matching for one-step visual generation.
\newblock \emph{arXiv preprint arXiv:2607.02375}, 2026.

\bibitem[Geng et~al.(2025)Geng, Deng, Bai, Kolter, and He]{geng2025meanflow}
Zhengyang Geng, Mingyang Deng, Xingjian Bai, J.~Zico Kolter, and Kaiming He.
\newblock Mean flows for one-step generative modeling.
\newblock \emph{arXiv preprint arXiv:2505.13447}, 2025.

\bibitem[Ghosh et~al.(2023)Ghosh, Hajishirzi, and Schmidt]{ghosh2023geneval}
Dhruba Ghosh, Hanna Hajishirzi, and Ludwig Schmidt.
\newblock {GenEval}: An object-focused framework for evaluating text-to-image
  alignment.
\newblock \emph{arXiv preprint arXiv:2310.11513}, 2023.

\bibitem[Gretton et~al.(2012)Gretton, Borgwardt, Rasch, Sch{\"o}lkopf, and
  Smola]{gretton2012mmd}
Arthur Gretton, Karsten~M. Borgwardt, Malte~J. Rasch, Bernhard Sch{\"o}lkopf,
  and Alexander Smola.
\newblock A kernel two-sample test.
\newblock \emph{Journal of Machine Learning Research}, 13\penalty0
  (25):\penalty0 723--773, 2012.

\bibitem[Gu et~al.(2024)Gu, Zhai, Zhang, Susskind, and
  Jaitly]{gu2024matryoshka}
Jiatao Gu, Shuangfei Zhai, Yizhe Zhang, Joshua Susskind, and Navdeep Jaitly.
\newblock Matryoshka diffusion models.
\newblock In \emph{International Conference on Learning Representations}, pages
  30234--30262, 2024.

\bibitem[He et~al.(2022)He, Chen, Xie, Li, Doll{\'a}r, and Girshick]{he2022mae}
Kaiming He, Xinlei Chen, Saining Xie, Yanghao Li, Piotr Doll{\'a}r, and Ross
  Girshick.
\newblock Masked autoencoders are scalable vision learners.
\newblock In \emph{Proceedings of the IEEE/CVF conference on computer vision
  and pattern recognition}, pages 16000--16009, 2022.

\bibitem[Ho et~al.(2020)Ho, Jain, and Abbeel]{ho2020ddpm}
Jonathan Ho, Ajay Jain, and Pieter Abbeel.
\newblock Denoising diffusion probabilistic models.
\newblock In \emph{Advances in Neural Information Processing Systems}, pages
  6840--6851, 2020.

\bibitem[Ho et~al.(2022)Ho, Saharia, Chan, Fleet, Norouzi, and
  Salimans]{ho2022cascaded}
Jonathan Ho, Chitwan Saharia, William Chan, David~J. Fleet, Mohammad Norouzi,
  and Tim Salimans.
\newblock Cascaded diffusion models for high fidelity image generation.
\newblock \emph{Journal of Machine Learning Research}, 23\penalty0
  (47):\penalty0 1--33, 2022.

\bibitem[Hoogeboom et~al.(2023)Hoogeboom, Heek, and
  Salimans]{hoogeboom2023simple}
Emiel Hoogeboom, Jonathan Heek, and Tim Salimans.
\newblock simple diffusion: End-to-end diffusion for high resolution images.
\newblock In \emph{International Conference on Machine Learning}, pages
  13213--13232, 2023.

\bibitem[Hoogeboom et~al.(2025)Hoogeboom, Mensink, Heek, Lamerigts, Gao, and
  Salimans]{sid2}
Emiel Hoogeboom, Thomas Mensink, Jonathan Heek, Kay Lamerigts, Ruiqi Gao, and
  Tim Salimans.
\newblock Simpler diffusion: 1.5 fid on imagenet512 with pixel-space diffusion.
\newblock In \emph{Proceedings of the Computer Vision and Pattern Recognition
  Conference}, pages 18062--18071, 2025.

\bibitem[Hu et~al.(2024)Hu, Wang, Fang, Fu, Cheng, and Yu]{hu2024ella}
Xiwei Hu, Rui Wang, Yixiao Fang, Bin Fu, Pei Cheng, and Gang Yu.
\newblock {ELLA}: Equip diffusion models with {LLM} for enhanced semantic
  alignment.
\newblock \emph{arXiv preprint arXiv:2403.05135}, 2024.

\bibitem[Jordan et~al.(2024)Jordan, Jin, Boza, You, Cesista, Newhouse, and
  Bernstein]{jordan2024muon}
Keller Jordan, Yuchen Jin, Vlado Boza, Jiacheng You, Franz Cesista, Laker
  Newhouse, and Jeremy Bernstein.
\newblock Muon: An optimizer for hidden layers in neural networks, 2024.

\bibitem[Karras et~al.(2017)Karras, Aila, Laine, and
  Lehtinen]{karras2018progressive}
Tero Karras, Timo Aila, Samuli Laine, and Jaakko Lehtinen.
\newblock Progressive growing of gans for improved quality, stability, and
  variation.
\newblock \emph{arXiv preprint arXiv:1710.10196}, 2017.

\bibitem[Karras et~al.(2024)Karras, Aittala, Kynk{\"a}{\"a}nniemi, Lehtinen,
  Aila, and Laine]{AG}
Tero Karras, Miika Aittala, Tuomas Kynk{\"a}{\"a}nniemi, Jaakko Lehtinen, Timo
  Aila, and Samuli Laine.
\newblock Guiding a diffusion model with a bad version of itself.
\newblock \emph{Advances in Neural Information Processing Systems},
  37:\penalty0 52996--53021, 2024.

\bibitem[Kingma and Welling(2014)]{kingma2014vae}
Diederik~P. Kingma and Max Welling.
\newblock Auto-encoding variational {Bayes}.
\newblock In \emph{International Conference on Learning Representations}, 2014.

\bibitem[Leng et~al.(2025)Leng, Singh, Hou, Xing, Xie, and Zheng]{repa-e}
Xingjian Leng, Jaskirat Singh, Yunzhong Hou, Zhenchang Xing, Saining Xie, and
  Liang Zheng.
\newblock Repa-e: Unlocking vae for end-to-end tuning of latent diffusion
  transformers.
\newblock In \emph{Proceedings of the IEEE/CVF International Conference on
  Computer Vision}, pages 18262--18272, 2025.

\bibitem[Leng et~al.(2026)Leng, Singh, Liang, Smith, Bell, Saha, Yuan, and
  Zheng]{leng2026diffusionbench}
Xingjian Leng, Jaskirat Singh, Zhanhao Liang, Ethan Smith, Martin Bell, Aninda
  Saha, Yuhui Yuan, and Liang Zheng.
\newblock Diffusionbench: On holistic evaluation of diffusion transformers.
\newblock \emph{arXiv preprint arXiv:2606.24888}, 2026.

\bibitem[Li and He(2025)]{li2025jit}
Tianhong Li and Kaiming He.
\newblock Back to basics: Let denoising generative models denoise.
\newblock \emph{arXiv preprint arXiv:2511.13720}, 2025.

\bibitem[Li et~al.(2024)Li, Katabi, and He]{li2024rcg}
Tianhong Li, Dina Katabi, and Kaiming He.
\newblock Return of unconditional generation: A self-supervised representation
  generation method.
\newblock In \emph{Advances in Neural Information Processing Systems}, 2024.

\bibitem[Li et~al.(2015)Li, Swersky, and Zemel]{li2015gmmn}
Yujia Li, Kevin Swersky, and Rich Zemel.
\newblock Generative moment matching networks.
\newblock In \emph{International conference on machine learning}, pages
  1718--1727. PMLR, 2015.

\bibitem[Lipman et~al.(2022)Lipman, Chen, Ben-Hamu, Nickel, and
  Le]{lipman2023flow}
Yaron Lipman, Ricky~TQ Chen, Heli Ben-Hamu, Maximilian Nickel, and Matt Le.
\newblock Flow matching for generative modeling.
\newblock \emph{arXiv preprint arXiv:2210.02747}, 2022.

\bibitem[Liu et~al.(2022)Liu, Gong, and Liu]{liu2023flow}
Xingchao Liu, Chengyue Gong, and Qiang Liu.
\newblock Flow straight and fast: Learning to generate and transfer data with
  rectified flow.
\newblock \emph{arXiv preprint arXiv:2209.03003}, 2022.

\bibitem[Ma et~al.(2024)Ma, Goldstein, Albergo, Boffi, Vanden-Eijnden, and
  Xie]{ma2024sit}
Nanye Ma, Mark Goldstein, Michael~S Albergo, Nicholas~M Boffi, Eric
  Vanden-Eijnden, and Saining Xie.
\newblock Sit: Exploring flow and diffusion-based generative models with
  scalable interpolant transformers.
\newblock In \emph{European Conference on Computer Vision}, pages 23--40.
  Springer, 2024.

\bibitem[Mahendran and Vedaldi(2015)]{mahendran2015understanding}
Aravindh Mahendran and Andrea Vedaldi.
\newblock Understanding deep image representations by inverting them.
\newblock In \emph{Proceedings of the IEEE Conference on Computer Vision and
  Pattern Recognition}, pages 5188--5196, 2015.

\bibitem[Oquab et~al.(2023)Oquab, Darcet, Moutakanni, Vo, Szafraniec, Khalidov,
  Fernandez, Haziza, Massa, El-Nouby, et~al.]{oquab2024dinov2}
Maxime Oquab, Timoth{\'e}e Darcet, Th{\'e}o Moutakanni, Huy Vo, Marc
  Szafraniec, Vasil Khalidov, Pierre Fernandez, Daniel Haziza, Francisco Massa,
  Alaaeldin El-Nouby, et~al.
\newblock Dinov2: Learning robust visual features without supervision.
\newblock \emph{arXiv preprint arXiv:2304.07193}, 2023.

\bibitem[Pan et~al.(2026)Pan, Chen, Shi, Hu, Yue, Wang, Wang, Li, and
  Yang]{pan2026rg}
Zihao Pan, Mengting Chen, Cheng Shi, Taihang Hu, Zhengrong Yue, Zikang Wang,
  Yuan Wang, Wei Li, and Sibei Yang.
\newblock Learning representations to guide pixel diffusion transformers.
\newblock \emph{arXiv preprint arXiv:2610.12978}, 2026.

\bibitem[Park et~al.(2019)Park, Liu, Wang, and Zhu]{park2019spade}
Taesung Park, Ming-Yu Liu, Ting-Chun Wang, and Jun-Yan Zhu.
\newblock Semantic image synthesis with spatially-adaptive normalization.
\newblock In \emph{Proceedings of the IEEE/CVF Conference on Computer Vision
  and Pattern Recognition}, pages 2337--2346, 2019.

\bibitem[Peebles and Xie(2023)]{peebles2023dit}
William Peebles and Saining Xie.
\newblock Scalable diffusion models with transformers.
\newblock In \emph{Proceedings of the IEEE/CVF international conference on
  computer vision}, pages 4195--4205, 2023.

\bibitem[Pernias et~al.(2024)Pernias, Rampas, Richter, Pal, and
  Aubreville]{pernias2024wuerstchen}
Pablo Pernias, Dominic Rampas, Mats~L. Richter, Christopher~J. Pal, and Marc
  Aubreville.
\newblock {W\"urstchen}: An efficient architecture for large-scale
  text-to-image diffusion models.
\newblock In \emph{International Conference on Learning Representations}, 2024.

\bibitem[Podell et~al.(2023)Podell, English, Lacey, Blattmann, Dockhorn,
  M{\"u}ller, Penna, and Rombach]{podell2023sdxl}
Dustin Podell, Zion English, Kyle Lacey, Andreas Blattmann, Tim Dockhorn, Jonas
  M{\"u}ller, Joe Penna, and Robin Rombach.
\newblock {SDXL}: Improving latent diffusion models for high-resolution image
  synthesis.
\newblock \emph{arXiv preprint arXiv:2307.01952}, 2023.

\bibitem[Ramesh et~al.(2022)Ramesh, Dhariwal, Nichol, Chu, and
  Chen]{ramesh2022unclip}
Aditya Ramesh, Prafulla Dhariwal, Alex Nichol, Casey Chu, and Mark Chen.
\newblock Hierarchical text-conditional image generation with {CLIP} latents.
\newblock \emph{arXiv preprint arXiv:2204.06125}, 2022.

\bibitem[Razavi et~al.(2019)Razavi, Van~den Oord, and
  Vinyals]{razavi2019vqvae2}
Ali Razavi, Aaron Van~den Oord, and Oriol Vinyals.
\newblock Generating diverse high-fidelity images with vq-vae-2.
\newblock \emph{Advances in neural information processing systems}, 32, 2019.

\bibitem[Rissanen et~al.(2023)Rissanen, Heinonen, and Solin]{rissanen2023ihdm}
Severi Rissanen, Markus Heinonen, and Arno Solin.
\newblock Generative modelling with inverse heat dissipation.
\newblock In \emph{International Conference on Learning Representations}, 2023.

\bibitem[Rombach et~al.(2022)Rombach, Blattmann, Lorenz, Esser, and
  Ommer]{rombach2022ldm}
Robin Rombach, Andreas Blattmann, Dominik Lorenz, Patrick Esser, and Bj{\"o}rn
  Ommer.
\newblock High-resolution image synthesis with latent diffusion models.
\newblock In \emph{Proceedings of the IEEE/CVF conference on computer vision
  and pattern recognition}, pages 10684--10695, 2022.

\bibitem[Salimans and Ho(2022)]{salimans2022progressive}
Tim Salimans and Jonathan Ho.
\newblock Progressive distillation for fast sampling of diffusion models.
\newblock \emph{arXiv preprint arXiv:2202.00512}, 2022.

\bibitem[Sim{\'e}oni et~al.(2025)Sim{\'e}oni, Vo, Seitzer, Baldassarre, Oquab,
  Jose, Khalidov, Szafraniec, Yi, Ramamonjisoa, et~al.]{simeoni2025dinov3}
Oriane Sim{\'e}oni, Huy~V. Vo, Maximilian Seitzer, Federico Baldassarre, Maxime
  Oquab, Cijo Jose, Vasil Khalidov, Marc Szafraniec, Seungeun Yi, Micha{\"e}l
  Ramamonjisoa, et~al.
\newblock {DINOv3}.
\newblock \emph{arXiv preprint arXiv:2508.10104}, 2025.

\bibitem[Singh et~al.(2026)Singh, Zheng, Wu, Zhang, Shechtman, and
  Xie]{singh2026raev2}
Jaskirat Singh, Boyang Zheng, Zongze Wu, Richard Zhang, Eli Shechtman, and
  Saining Xie.
\newblock Improved baselines with representation autoencoders.
\newblock \emph{arXiv preprint arXiv:2605.18324}, 2026.

\bibitem[Sohl-Dickstein et~al.(2015)Sohl-Dickstein, Weiss, Maheswaranathan, and
  Ganguli]{sohl2015deep}
Jascha Sohl-Dickstein, Eric Weiss, Niru Maheswaranathan, and Surya Ganguli.
\newblock Deep unsupervised learning using nonequilibrium thermodynamics.
\newblock In \emph{International Conference on Machine Learning}, pages
  2256--2265, 2015.

\bibitem[Song et~al.(2020)Song, Sohl-Dickstein, Kingma, Kumar, Ermon, and
  Poole]{song2021score}
Yang Song, Jascha Sohl-Dickstein, Diederik~P Kingma, Abhishek Kumar, Stefano
  Ermon, and Ben Poole.
\newblock Score-based generative modeling through stochastic differential
  equations.
\newblock \emph{arXiv preprint arXiv:2011.13456}, 2020.

\bibitem[Song et~al.(2023)Song, Dhariwal, Chen, and
  Sutskever]{song2023consistency}
Yang Song, Prafulla Dhariwal, Mark Chen, and Ilya Sutskever.
\newblock Consistency models.
\newblock In \emph{International Conference on Machine Learning}, pages
  32211--32252, 2023.

\bibitem[Tian et~al.(2024)Tian, Jiang, Yuan, Peng, and Wang]{VAR}
Keyu Tian, Yi Jiang, Zehuan Yuan, Bingyue Peng, and Liwei Wang.
\newblock Visual autoregressive modeling: Scalable image generation via
  next-scale prediction.
\newblock \emph{Advances in neural information processing systems},
  37:\penalty0 84839--84865, 2024.

\bibitem[van~den Oord et~al.(2017)van~den Oord, Vinyals, and
  Kavukcuoglu]{vandenoord2017vqvae}
Aaron van~den Oord, Oriol Vinyals, and Koray Kavukcuoglu.
\newblock Neural discrete representation learning.
\newblock In \emph{Advances in Neural Information Processing Systems}, 2017.

\bibitem[Wang et~al.(2025{\natexlab{a}})Wang, Gao, Zhu, Huang, and
  Wang]{pixnerd}
Shuai Wang, Ziteng Gao, Chenhui Zhu, Weilin Huang, and Limin Wang.
\newblock Pixnerd: Pixel neural field diffusion.
\newblock \emph{arXiv preprint arXiv:2507.23268}, 2025{\natexlab{a}}.

\bibitem[Wang et~al.(2025{\natexlab{b}})Wang, Tian, Huang, and Wang]{ddt}
Shuai Wang, Zhi Tian, Weilin Huang, and Limin Wang.
\newblock Ddt: Decoupled diffusion transformer.
\newblock \emph{arXiv preprint arXiv:2504.05741}, 2025{\natexlab{b}}.

\bibitem[Yang et~al.(2026{\natexlab{a}})Yang, Geng, Ju, Tian, and
  Wang]{yang2026fdloss}
Jiawei Yang, Zhengyang Geng, Xuan Ju, Yonglong Tian, and Yue Wang.
\newblock Representation {Fr\'{e}chet} loss for visual generation.
\newblock \emph{arXiv preprint arXiv:2604.28190}, 2026{\natexlab{a}}.

\bibitem[Yang et~al.(2026{\natexlab{b}})Yang, Li, Fan, Tian, and
  Wang]{yang2025detok}
Jiawei Yang, Tianhong Li, Lijie Fan, Yonglong Tian, and Yue Wang.
\newblock Latent denoising makes good tokenizers.
\newblock In \emph{International Conference on Learning Representations}, pages
  61642--61668, 2026{\natexlab{b}}.

\bibitem[Yao et~al.(2025)Yao, Yang, and Wang]{yao2025vavae}
Jingfeng Yao, Bin Yang, and Xinggang Wang.
\newblock Reconstruction vs. generation: Taming optimization dilemma in latent
  diffusion models.
\newblock In \emph{Proceedings of the Computer Vision and Pattern Recognition
  Conference}, pages 15703--15712, 2025.

\bibitem[Yin et~al.(2024)Yin, Gharbi, Zhang, Shechtman, Durand, Freeman, and
  Park]{yin2024dmd}
Tianwei Yin, Micha{\"e}l Gharbi, Richard Zhang, Eli Shechtman, Fr{\'e}do
  Durand, William~T. Freeman, and Taesung Park.
\newblock One-step diffusion with distribution matching distillation.
\newblock In \emph{Proceedings of the IEEE/CVF Conference on Computer Vision
  and Pattern Recognition}, pages 6613--6623, 2024.

\bibitem[Yu et~al.(2024)Yu, Kwak, Jang, Jeong, Huang, Shin, and
  Xie]{yu2025repa}
Sihyun Yu, Sangkyung Kwak, Huiwon Jang, Jongheon Jeong, Jonathan Huang, Jinwoo
  Shin, and Saining Xie.
\newblock Representation alignment for generation: Training diffusion
  transformers is easier than you think.
\newblock \emph{arXiv preprint arXiv:2410.06940}, 2024.

\bibitem[Yue et~al.(2026)Yue, Zhang, Zeng, Chen, Wang, Zhuang, Dong, Wang,
  Wang, and Wang]{yue2025uniflow}
Zhengrong Yue, Haiyu Zhang, Xiangyu Zeng, Boyu Chen, Chenting Wang, Shaobin
  Zhuang, Lu Dong, Yi Wang, Limin Wang, and Yali Wang.
\newblock Uniflow: A unified pixel flow tokenizer for visual understanding and
  generation.
\newblock In \emph{International Conference on Learning Representations}, pages
  84741--84772, 2026.

\bibitem[Zhang et~al.(2023)Zhang, Rao, and Agrawala]{zhang2023controlnet}
Lvmin Zhang, Anyi Rao, and Maneesh Agrawala.
\newblock Adding conditional control to text-to-image diffusion models.
\newblock In \emph{Proceedings of the IEEE/CVF International Conference on
  Computer Vision}, pages 3836--3847, 2023.

\bibitem[Zheng et~al.(2025)Zheng, Ma, Tong, and Xie]{zheng2025rae}
Boyang Zheng, Nanye Ma, Shengbang Tong, and Saining Xie.
\newblock Diffusion transformers with representation autoencoders.
\newblock \emph{arXiv preprint arXiv:2510.11690}, 2025.

\bibitem[Zhou et~al.(2025)Zhou, Li, Hu, Chen, and Gu]{ig}
Xingyu Zhou, Qifan Li, Xiaobin Hu, Hai Chen, and Shuhang Gu.
\newblock Guiding a diffusion transformer with the internal dynamics of itself.
\newblock \emph{arXiv preprint arXiv:2512.24176}, 2025.

\end{thebibliography}

}
\end{document}